\documentclass[11pt,a4paper]{article}

\usepackage[utf8]{inputenc}
\usepackage[T1]{fontenc}
\usepackage{newpxtext,newpxmath} %
\usepackage{microtype}         %
\usepackage{url}
\usepackage{algorithm}
\usepackage{algpseudocode} 
\usepackage{booktabs}
\usepackage{subcaption}
\usepackage[bottom]{footmisc}
\usepackage{caption}
\usepackage{float}
\usepackage{amsmath}
\usepackage{bm}
\usepackage{makecell}
\usepackage{appendix}
\usepackage{titlesec}
\usepackage{xparse}
\usepackage{mathtools}

\usepackage{hyperref}
\usepackage{cleveref}  %
\hypersetup{
    colorlinks=true,
    linkcolor=blue,
    filecolor=magenta,      
    urlcolor=cyan,
    pdftitle={Overleaf Example},
    pdfpagemode=FullScreen,
    }

\AtBeginEnvironment{appendices}{\crefalias{section}{appendix}}
\usepackage{geometry}
\usepackage{parskip}           %

\usepackage{xcolor}
\definecolor{abstractbg}{RGB}{235,243,250} %
\definecolor{titlecolor}{RGB}{20,20,40}      %

\usepackage{tcolorbox}
\tcbset{
  fullwidthabstract/.style={
    width=\textwidth,
    colback=abstractbg,
    colframe=abstractbg,
    boxrule=0pt,
    arc=4pt,
    left=10pt,
    right=10pt,
    top=10pt,
    bottom=10pt,
  },
}

\algtext*{EndFunction}
\algtext*{EndIf}
\algtext*{EndFor}
\algtext*{EndWhile}

\usepackage{siunitx}
\usepackage{multirow}
\usepackage{longtable}
\usepackage{seqsplit}

\crefname{figure}{Figure}{Figures}
\Crefname{figure}{Figure}{Figures}
\crefname{table}{Table}{Tables}
\Crefname{table}{Table}{Tables}
\crefname{algorithm}{Algorithm}{Algorithms}
\Crefname{algorithm}{Algorithm}{Algorithms}
\crefname{section}{Section}{Sections}
\Crefname{section}{Section}{Sections}
\crefname{equation}{Equation}{Equations}
\Crefname{equation}{Equation}{Equations}
\crefname{appendix}{Appendix}{Appendix}
\Crefname{appendix}{Appendix}{Appendix}

\newcommand{\companylogo}{\includegraphics[height=1.5cm]{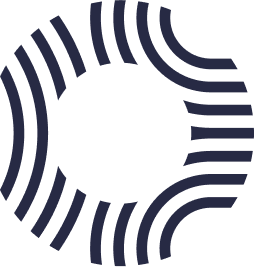}}

\newcommand{\modelname}{Mofasa}
\newcommand{\dbname}{\texttt{\modelname DB}}

\usepackage[
natbib=true,
backend=biber,
style=numeric,
sorting=ynt,
maxbibnames=100,
backref=true,
maxcitenames=2,
]{biblatex}
\DeclareFieldFormat*{title}{\mkbibemph{#1}}
\AtEveryBibitem{\clearfield{month}}
\AtEveryCitekey{\clearfield{month}}

\DeclareMathOperator*{\argmax}{arg\,max}

\newcommand{\norm}[1]{\left\lVert#1\right\rVert}
\DeclarePairedDelimiterX{\infdivx}[2]{(}{)}{#1\;\delimsize\|\;#2}
\newcommand{\KLD}[2]{D_{\text{KL}}\infdivx{#1}{#2}}

\newcommand{\realdomain}{\mathbb{R}}
\newcommand{\posdomain}{\mathbb{R}_{>0}}

\newcommand{\atomdomain}{\mathbb{A}}
\newcommand{\unitdomain}{[0, 1)}
\newcommand{\atom}{a}
\newcommand{\atoms}{\bm{A}}

\newcommand{\fposition}{\bm{f}}
\newcommand{\fpositions}{\bm{F}}
\newcommand{\lattice}{\bm{L}}
\newcommand{\system}{\bm{S}}
\newcommand{\latents}{\bm{Z}}
\newcommand{\latent}{\bm{z}}
\newcommand{\llatents}{\latents_{\text{L}}}

\newcommand{\glatents}{\latents_{\text{G}}}

\newcommand{\enc}{\textsc{Enc}}
\newcommand{\dec}{\textsc{Dec}}
\newcommand{\bottleneck}{\textsc{Bottleneck}}
\newcommand{\loss}{\mathcal{L}}
\newcommand{\angstrom}{\mbox{\normalfont\AA}}

\newcommand{\thetab}{\bm{\theta}}

\newcommand{\vb}{\bm{v}}
\newcommand{\baralpha}{\bar \alpha}
\newcommand{\node}{\bm{v}}
\newcommand{\nodes}{\bm{V}}
\newcommand{\edge}{\bm{e}}
\newcommand{\edges}{\bm{E}}
\newcommand{\nn}{\phi}  %
\newcommand{\nnn}[1]{\nn^{\text{#1}}_{\text{N}}}
\newcommand{\lnnn}[1]{\nn^{\text{#1}}_{\text{N,L}}}
\newcommand{\gnnn}[1]{\nn^{\text{#1}}_{\text{N,G}}}
\newcommand{\enn}[1]{\nn^{\text{#1}}_{\text{E}}}
\newcommand{\lenn}[1]{\nn^{\text{#1}}_{\text{E,L}}}
\newcommand{\genn}[1]{\nn^{\text{#1}}_{\text{E,G}}}
\newcommand{\eps}{\bm{\epsilon}}
\newcommand{\m}{\bm{m}}

\begin{document}

\vspace*{-2cm}
\begin{tcolorbox}[fullwidthabstract, title={\vspace{3mm} \hspace{-4cm}\makebox[0pt][l]{\companylogo}\hspace{3.5cm} \centering \color{titlecolor}{ \fontsize{18}{20}\selectfont \emph{\modelname}: \fontsize{16}{20}\selectfont A Step Change in \\ Metal-Organic Framework Generation}}]

\begin{center}
Vaidotas Šimkus, Anders Christensen, Steven Bennett, Ian Johnson,\\ Mark Neumann, James Gin, Jonathan Godwin, Benjamin Rhodes\\[1em]
\texttt{\{vaidas,ben\}@orbitalindustries.com} \\
Orbital \\[1.5em]

\end{center}
\modelname{} is an all-atom latent diffusion model with state-of-the-art performance for generating Metal-Organic Frameworks (MOFs). These are highly porous crystalline materials used to harvest water from desert air, capture carbon dioxide, store toxic gases and catalyse chemical reactions. In recognition of their value, the development of MOFs recently received a Nobel Prize in Chemistry. \\

In many ways, MOFs are 
well-suited
for exploiting generative models in chemistry: they are rationally-designable materials with a large combinatorial design space and strong structure-property couplings. And yet, to date, a high performance generative model has been lacking. To fill this gap, we introduce \modelname{}, a general-purpose latent diffusion model that jointly samples positions, atom-types and lattice vectors for systems as large as 500 atoms. \modelname{} avoids handcrafted assembly algorithms common in the literature, unlocking the simultaneous discovery of metal nodes, linkers and topologies. \\

To help the scientific community build on our work, we release \dbname, an annotated library of hundreds of thousands of sampled MOF structures, along with a user-friendly web interface for search and discovery: \url{https://mofux.ai/}.

\end{tcolorbox}

\vspace{1.5em}

Discovering novel periodic crystal structures at scale is a grand challenge in materials science. 
In few places is this opportunity greater than with Metal-Organic Frameworks (MOFs)---highly porous materials critical for next-generation climate technologies \citep{nobel2025chemistry}.

Traditionally, MOFs are viewed as modular compositions of metal nodes and organic linkers arranged in a specific network topology. 
While new materials can be found by traversing the combinatorial space of known building blocks, discovering truly novel chemistry requires more than just recombining existing components---we must adopt an \emph{all-atom} design approach.

All-atom generative models present a promising path forward to discover truly novel chemistries due to their unconstrained nature \cite{xieCrystalDiffusionVariational2022,zeniMatterGenGenerativeModel2025,joshiAllatomDiffusionTransformers2025}.
Yet, they have failed, so far, to scale to large crystal systems containing many atoms in the unit cell. 
This is odd, because generative models of text, images, and---most pertinently---biomolecules scale well as a function of dimensionality \cite{watsonNovoDesignProtein2023, abramsonAccurateStructurePrediction2024}.

This motivates us to find a scalable recipe for all-atom generation of large crystal structures.
Our method takes ADiT \citep{joshiAllatomDiffusionTransformers2025} as an initial template, and re-engineers the architecture, parameterization, training loss, and sampling algorithm. In this sense, our approach is proudly incremental: we eschew methodological novelty in favor of thorough engineering. However, once implemented correctly, the \emph{results} are far from incremental, and it is these results that constitute the bulk of this report.

\begin{figure}[h]
    \centering
    \includegraphics[width=1.0\linewidth]{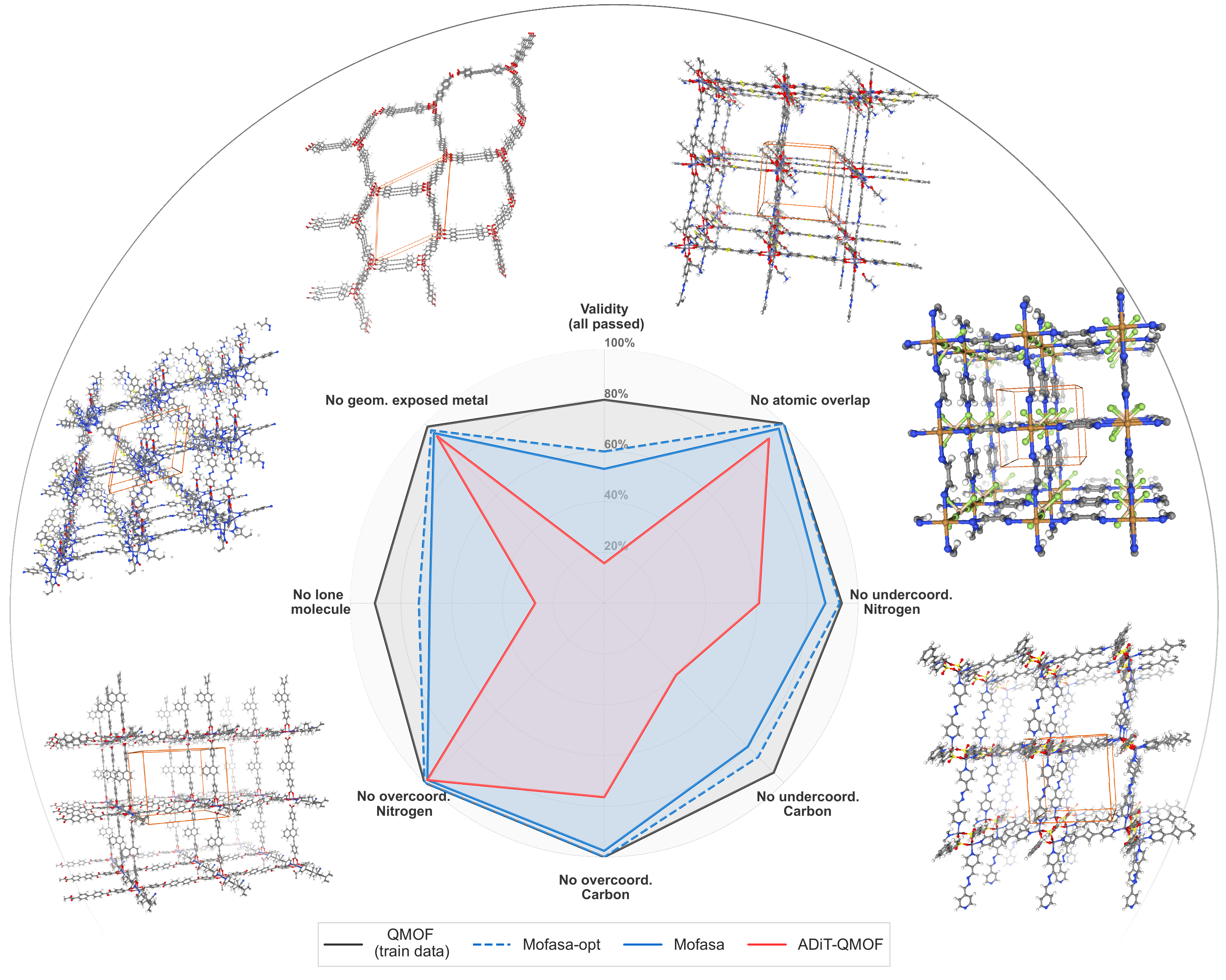}
    \caption{\textbf{Validation of geometric structure with MOFChecker}. \modelname{} demonstrates a step-change improvement over the leading all-atom baseline, ADiT \cite{joshiAllatomDiffusionTransformers2025}, increasing overall MOFChecker validity by $3.8{\times}$ from 15.7\% to 59.9\%.}
    \label{fig:mofchecker-radar}
\end{figure}
\begin{figure}[h]
    \vspace{-2cm}
    \centering
    \includegraphics[width=0.9\linewidth]{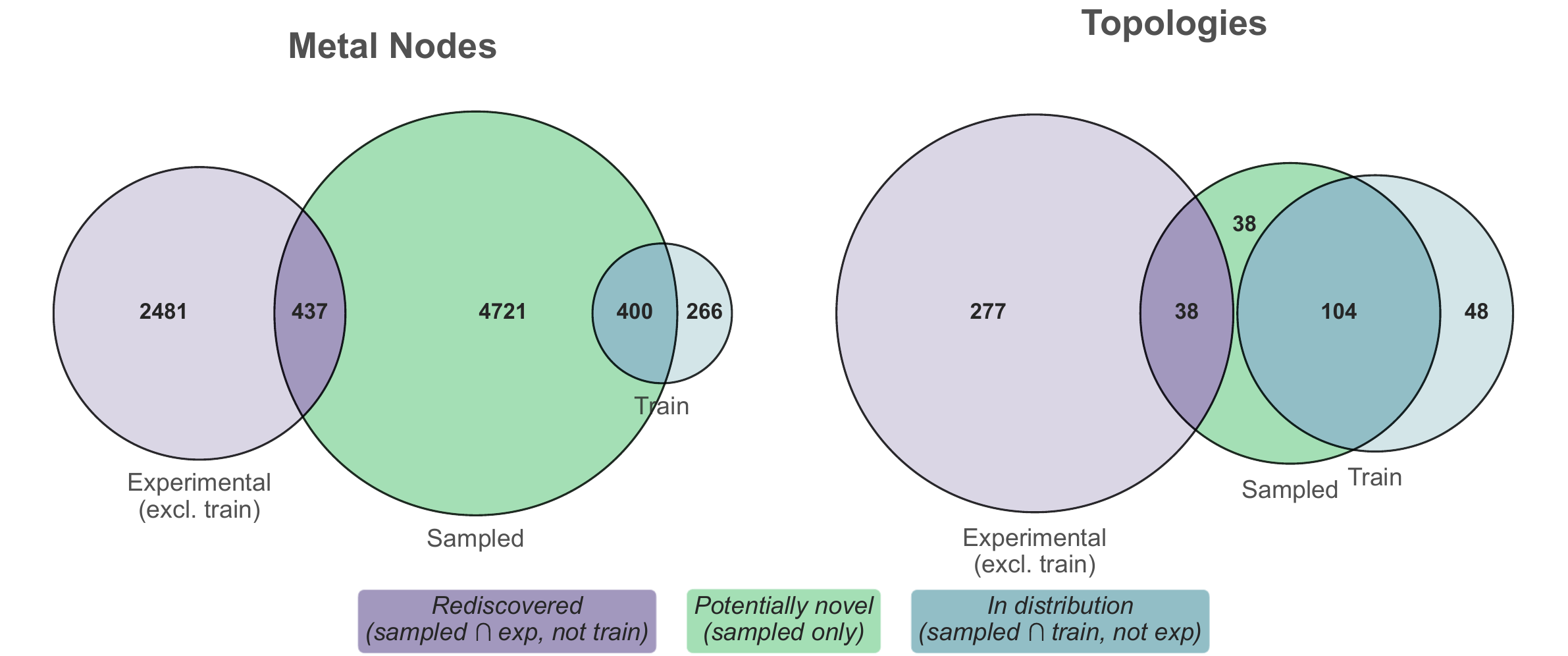}
    \caption{\textbf{Rediscovery and novelty analysis.} \modelname{} demonstrates strong generalization by \emph{rediscovering} 437 nodes + 38 topologies absent from the training set. Beyond generating known chemistry, the model also generates MOFs with chemistries unseen in experimental databases.}
    \label{fig:venn-rediscovery}
\end{figure}

\clearpage
\vspace*{-12mm}
Our main contributions are:
\begin{itemize}
\item High-fidelity generation (53--62\% \texttt{MOFChecker} validity\footnote{Exact percentage varies with training set and whether samples are relaxed with an MLIP.}) of entire 3D crystal structures (positions, atom-types, lattice) for MOFs up to 500 atoms. On QMOF, we push SoTA forward by $>3 \times$ (see \cref{fig:mofchecker-radar}). 
\item From a total of 100k samples, 40--46\% are \emph{valid, novel and unique} as measured by \texttt{MOFid} + \texttt{MOFChecker}. Importantly, the training set (QMOF) imposes an upper limit of $70\%$.
\item A \emph{rediscovery rate} of $8.5\%$ and $50.0\%$ for nodes and topologies, respectively. \modelname{} rediscovers a significant number of metal nodes and topologies that are not present in the training data, but are present in other experimental databases.
\item An open source database, \dbname{}, containing ${\sim}200$k generated structures annotated with a large set of descriptors and properties to enable screening by composition, types of nodes and linkers, topology, space group, porosity and more (see \cref{sec:mofasadb}).
\item A scalable data-agnostic architecture. Without modification, \modelname{} can be jointly trained on MOFs, other porous materials classes, organic molecular crystals, and gas-phase organics + transition metal complexes.
\end{itemize}

\section{Results}
\label{sec:results}

\subsection{Comparison to state-of-the-art}

A unified benchmark for MOF generation is lacking. As a consequence, many recent works \citep{inizan2025system, parkMultimodalConditionalDiffusion2025, yanMOFADiscoveringMaterials2025} use disparate metrics and datasets, rendering direct comparison infeasible.
Nonetheless, we believe ADiT \citep{joshiAllatomDiffusionTransformers2025}, MOFFlow-2 \citep{kimFlexibleMOFGeneration2025} and MOF-BFN \citep{jiaoMOFBFNMetalOrganicFrameworks2025} are a fair representation of the state-of-the-art. 
All these works report \texttt{MOFChecker} validity \cite{jin2025mofchecker}, a rigorous metric that assesses structural and compositional validity of MOFs by penalizing flaws such as floating atoms and incorrect coordination (see \cref{tab:mofchecker-qmof,tab:mofchecker-expmof} for a full breakdown of the criteria).
\vspace{1mm}

\begin{table}[b!]
    \centering
    \begin{tabular}{@{}r@{}lccclcc@{}}
        \toprule
        &\textbf{Model} & \multirow{2}{6em}{\centering{\textbf{Sample count ($n$)}}} & \textbf{Validity (\%)} && \multicolumn{3}{c}{\textbf{Training dataset}} \\
        \cmidrule{6-8}
        & & & && \textbf{Name} & \textbf{Count ($n$)} & \textbf{Validity (\%)} \\ 
        \midrule
        $\dagger$   & ADiT & 1k & 15.7 && QMOF \cite{rosenMachineLearningQuantumchemical2021} & 14k & 80.0\\
                    & \textbf{Mofasa} & 202k & 52.9 && QMOF \cite{rosenMachineLearningQuantumchemical2021} & 14k & 80.0\\
                    & \textbf{Mofasa-opt} & 202k & \textbf{59.9} && QMOF \cite{rosenMachineLearningQuantumchemical2021} & 14k & 80.0\\
        \midrule
        $\dagger$   & MOFFlow-2 & 10k & 38.8 && BW \cite{boydComputationalDevelopmentNanoporous2017,boydDatadrivenDesignMetal2019} & 157k & 100.0 \\
        $\dagger$   & MOF-BFN & 1k & 32.3 && BW \cite{boydComputationalDevelopmentNanoporous2017,boydDatadrivenDesignMetal2019} & 324k & --\\
                    &\textbf{Mofasa} & 10k & \textbf{44.8} && BW \cite{burnerARCMOFDiverseDatabase2023} & 250k & 85.8 \\
        \midrule
                    & \textbf{Mofasa} & 250k & 52.2 && Exp. \cite{rosenMachineLearningQuantumchemical2021,burnerARCMOFDiverseDatabase2023,zhaoCoREMOFDB2025} & 49k & 84.2 \\
                    & \textbf{Mofasa-opt} & 225k & \textbf{62.8} && Exp.\ \cite{rosenMachineLearningQuantumchemical2021,burnerARCMOFDiverseDatabase2023,zhaoCoREMOFDB2025} & 49k & 84.2 \\
        \bottomrule
    \end{tabular}
    \caption{\textbf{Comparison of MOFChecker validity (\%) across datasets.} $\dagger$ indicates baseline models, and \textbf{bold} (Mofasa/Ours). Right hand side shows the training dataset and its statistics for each model. In this table and elsewhere, Experimental (Exp.) = QMOF \cite{rosenMachineLearningQuantumchemical2021} + CoRE-MOF-2024 (computation-ready splits) \cite{zhaoCoREMOFDB2025} + ARC-MOF DB12 \& DB14 \cite{burnerARCMOFDiverseDatabase2023}. See more details in \cref{sec: training data}.}
    \label{tab:mof_validity}
\end{table}
\texttt{MOFChecker} validity scores for \modelname{} (trained on different datasets) are shown in \cref{tab:mof_validity}, along with reference values for the training data and baseline methods. \modelname{-opt} refers to model samples that have been geometry-optimized with an MLIP (see \cref{sec:geoopt}).

\begin{figure}[t]
    \vspace*{-8mm}
    \centering
    \includegraphics[width=0.9\linewidth]{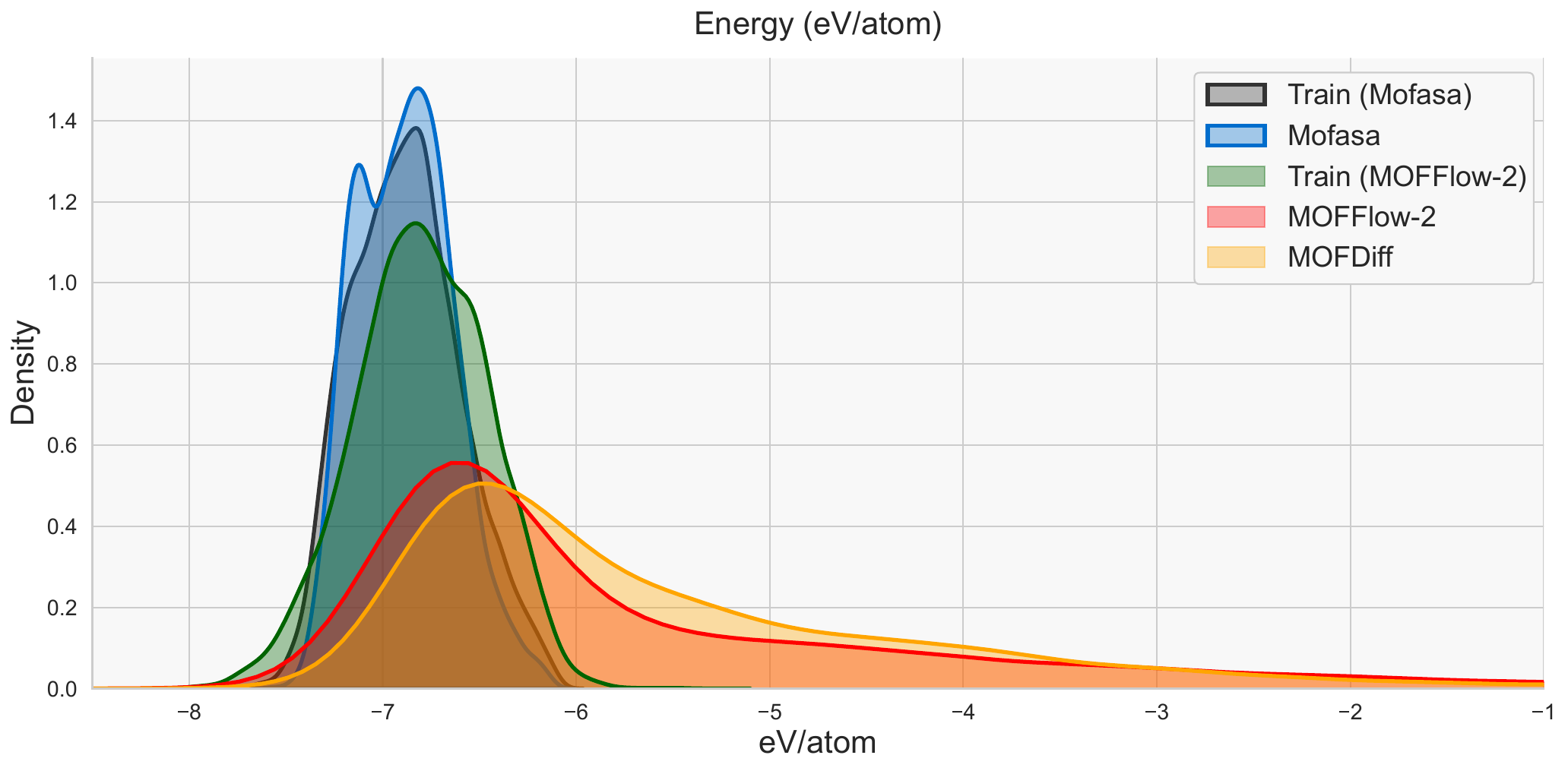}
    \caption{\textbf{Potential energy histograms on the Boyd-Woo (BW) \citep{boydComputationalDevelopmentNanoporous2017,boydDatadrivenDesignMetal2019} dataset.}
    Note that, \modelname{} (black) and MOFFlow-2 (red) are trained on slightly different subsets of BW (see \cref{sec: training data}) and the \modelname{} energies (black, blue) were computed with \texttt{Orb-v3-con-inf-omat}\citep{rhodesOrbv3AtomisticSimulation2025} vs UMA \citep{wood2025family} for the MOFFlow-2 and MOFDiff. Nonetheless, the trend is clear: \modelname{} is much better at matching the energy distribution of its training data.
    }
    \label{fig:energy-density-arcmof0}
\end{figure}

One significant limitation of \texttt{MOFChecker} is that it is discrete and brittle. Small, structural deviations can cause physically reasonable structures to be flagged as invalid. 
One way to gauge the magnitude of structural errors is to study the energies of the samples using MLIPs, which are fast DFT surrogate models. In \cref{fig:energy-density-arcmof0}, we plot the energy per atom of \modelname{} samples (and its training set) against the samples and training set of MOFFlow-2 and MOFDiff \citep{fuMOFDiffCoarsegrainedDiffusion2024}. It is evident that \modelname{} is substantially better at following the true distribution of energies. 
Interestingly, this performance gap is not clear from the \texttt{MOFChecker} scores in \cref{tab:mof_validity}, where \modelname{}'s $44.8\%$ validity is only modestly higher than MOFFlow-2's $38.8\%$.
This discrepancy underscores the need for distributional coverage evaluations alongside geometric checks.

\subsection{Validity, novelty, and uniqueness}

Validity scores and distributional similarities can be trivially maximized by a model that strictly memorizes the training data.
To assess generalization, standard practice is to compute a Valid, Novel, and Unique (VNU) score.
Here, a sample is considered \emph{novel} if it does not match any data point in the training set, and \emph{unique} if it does not match any other sample in a fixed set of generated samples. What, however, should count as a `match'? 
This is a matter of an ongoing debate with many pitfalls \citep{negishiContinuousUniquenessNovelty2025, juelsholtContinuedChallengesHighThroughput2025}. We do not claim to resolve all such pitfalls here and our definition of ``novelty'' should be interpreted with care.

We leverage prior efforts to uniquely identify MOFs using \texttt{MOFid} \citep{buciorIdentificationSchemesMetal2019}. \texttt{MOFid} decomposes a 3D MOF into building blocks: metal nodes, organic linkers, network topology, and catenation. The final identifier is a string representation of these four components that is both convenient and chemically-informed. 
One drawback is incomplete coverage: \texttt{MOFid} only succeeds for ${\sim}85\%$ of systems in QMOF, often failing on chemically valid structures such as 2D or rod-like MOFs \citep[such as MOF-74, see][]{buciorIdentificationSchemesMetal2019} and is sensitive to geometric perturbations (see \cref{tab:topology_change_percentage_breakdown,tab:topology_stability,tab:topology_top15_changes}).

\begin{figure}[h]
    \centering
    \makebox[\linewidth][c]{%
        \hspace*{2cm}%
        \includegraphics[width=\linewidth]{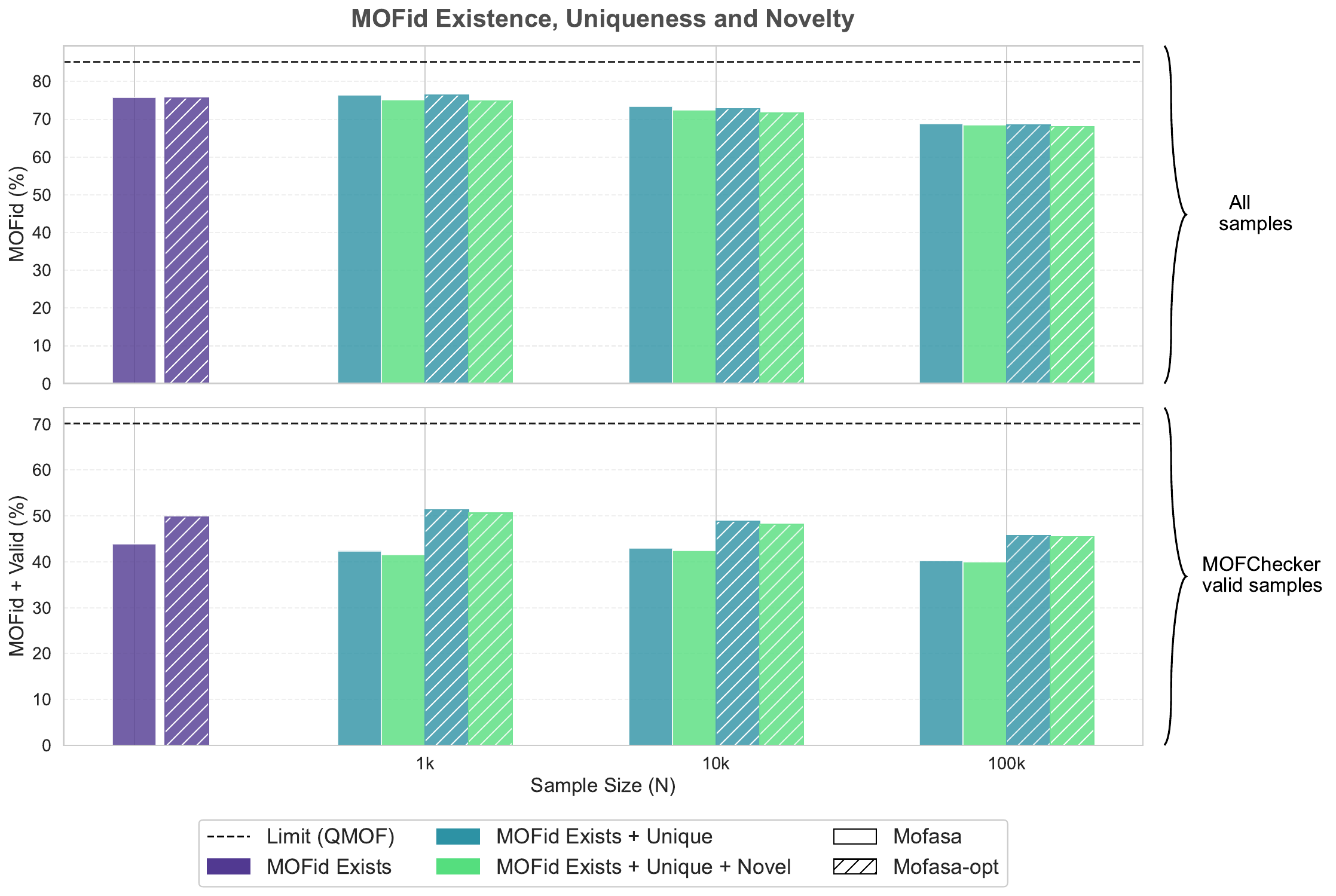}%
    }
    \caption{\textbf{Validity, Novelty, and Uniqueness (VNU) analysis.} Percentage of QMOF samples for which MOFids exists (purple) and is unique (teal) and is novel (light green). \textbf{Top row:} all samples without validity constraints; the maximum score is $84\%$. \textbf{Bottom row:} only MOFChecker valid systems; the maximum score is $70\%$. See Appendix Table \ref{tab:vnu} for full numerical results.}
    \label{fig:vnu}
\end{figure}

\Cref{fig:vnu} shows \modelname{'s} ``success rates'' for a range of sample sizes and different combinations of VNU conditions. Importantly, at 100,000 samples, we see that:
\begin{itemize}
    \item \modelname{} obtains ${\sim}68\%$ (maximum ${\sim}84\%$) NU score (\texttt{MOFid exists + novel + unique}).
    \item \modelname{}(-opt) obtains $40\%$ ($46\%$) (maximum ${\sim}70\%$) VNU score (NU + \texttt{MOFChecker-valid}).
\end{itemize}

\subsection{Rediscovery of unseen experimental nodes, linkers, and topologies}

MOFs can be computationally designed without machine learning \citep{moghadam2024progress} using rule-based systems for assembling libraries of building blocks according to predefined topologies \citep{leeComputationalScreeningTrillions2021}. 
While these rule-based systems can be combined with generative models of building blocks \citep{duanBuildingBlockAwareGenerative2025, yanMOFADiscoveringMaterials2025}, they remain fundamentally constrained. 
The presupposition of a known topology introduces a degree of subjectivity \citep{glasby2024topological}, hinders the exploration of polymorphs \citep{xuExperimentallyValidatedInitio2023}, and creates a barrier to incorporating new sources of non-porous systems into the training data.

In contrast, \modelname{} generates atoms directly in the full 3D crystal. Topologies and building blocks are not presupposed, they \emph{emerge} from the generative process. 
This allows us to simultaneously discover new MOF ``components'' (nodes, linkers and topologies) as demonstrated in \cref{table: component rediscovery}.
The key figure in this table is the \emph{rediscovery rate}, which is the fraction of unique and novel sampled components that exist in an experimental database, but \emph{not} in the training data. See \cref{fig:venn-rediscovery} for a visual summary. 
Metal nodes have a rediscovery rate of $8.5\%$ and topologies have a rediscovery rate of $50.0\%$, which proves \modelname{} performs useful generalization. The rediscovery rate for linkers is very low ($0.5\%$) which is expected given the enormous chemical diversity of possible linkers compared to the more constrained set of stable metal nodes and topologies.

\begin{table}[t]
\centering
\caption{\textbf{Analysis of unique MOF Components.} For each data source (experimental, training set, samples) we count the numbers of unique components. We then remove all components in the training set from the sampled components, to obtain unique + novel components. Finally, a unique and novel component is counted as \emph{Rediscovered} if it exists in an experimental source.}
\label{table: component rediscovery}
\begin{tabular}{@{}lrrrrr@{}}
\toprule
\textbf{Component} &
\makecell[c]{\textbf{\# Unique}\\\textbf{(Experiment)}} &
\makecell[c]{\textbf{\# Unique}\\\textbf{(Train)}} &
\makecell[c]{\textbf{\# Unique}\\\textbf{(Samples)}} &
\makecell[c]{\textbf{\# Unique + Novel}\\ \textbf{(Samples)}} &
\makecell[c]{\textbf{\# Rediscovered}\\ \textbf{(Rate \%)}} \\
\midrule
MOFid & 34,873 & 11,014 & 134,800 & 134,351 & 182 (0.1\%) \\
Nodes & 3,584 & 666 & 5,558 & 5,158 & 437 (8.5\%) \\
Linkers & 15,188 & 6,302 & 111,047 & 109,831 & 538 (0.5\%) \\
Linker combos & 10,682 & 3,911 & 63,128 & 62,996 & 92 (0.1\%) \\
Topologies & 467 & 152 & 180 & 76 & 38 (50.0\%) \\
\bottomrule
\end{tabular}
\end{table}

\subsection{Distributional realism of simple properties}
\begin{figure}[h!]
    \centering
    \includegraphics[width=0.98\linewidth]{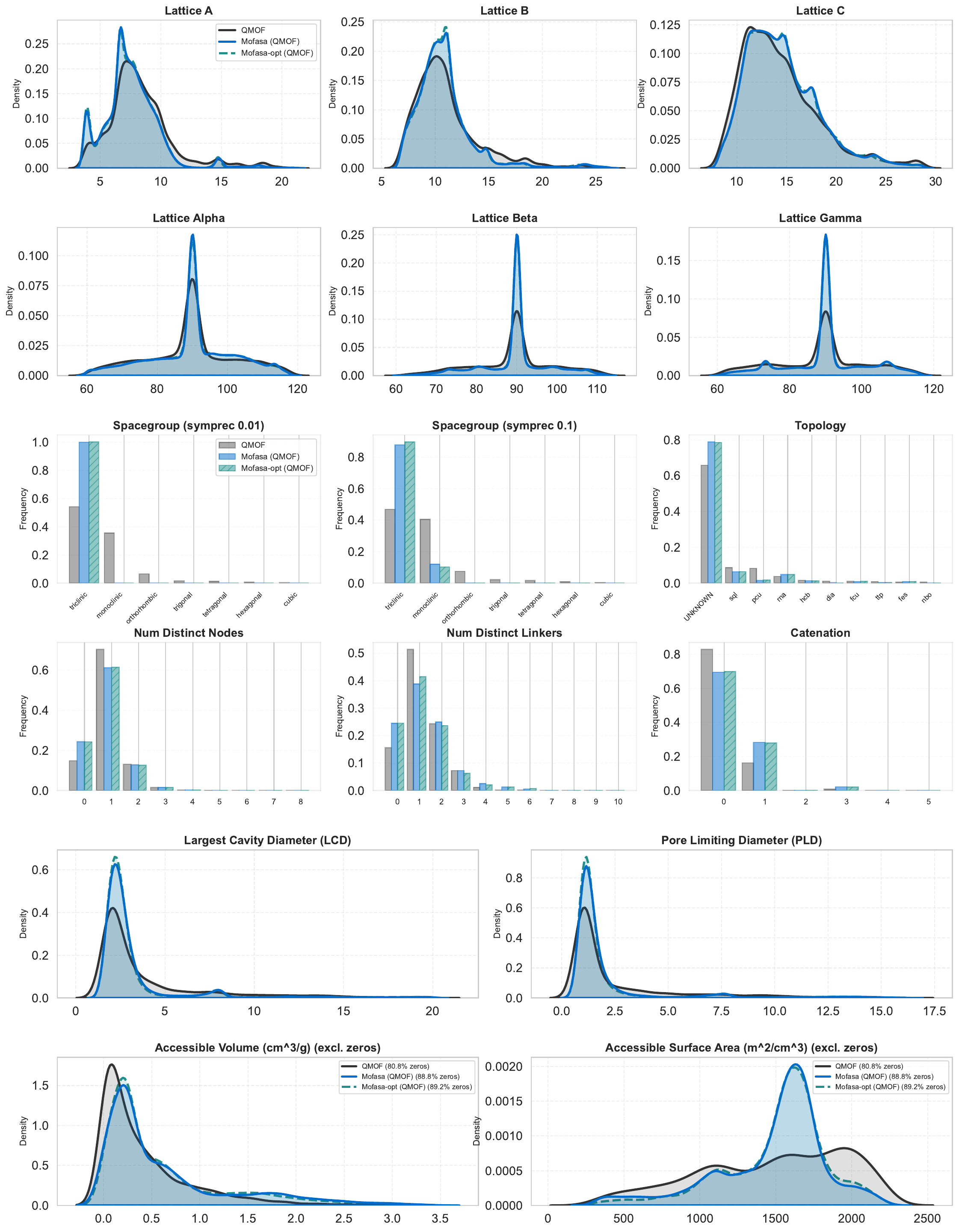}
    \caption{\textbf{Marginal distributions of simple properties of real data (QMOF, black) and generated samples (\modelname{}, blue).} There is a significant amount of distributional overlap for all properties, with no signs of severe mode collapse. Nonetheless, several areas remain for improvement: generated MOFs are on average, slightly too small, too cubic, far too likely to be triclinic, more likely to fail \texttt{MOFid} identification (resulting in 0 nodes/linkers) and insufficiently porous as assessed by Zeo++ with a $1.86$\angstrom{} Nitrogen probe (note the legends of the final row, which state the percentage of systems for which zero volume/area is accessible).}
    \label{fig:property-histograms}
\end{figure}

\Cref{fig:property-histograms} shows a range of simple property histograms for \dbname{} (${\sim}200$k samples from our QMOF model). The top row displays the distributions of lattice vector lengths ($a, b, c$), which are of ascending order in the Niggli-reduced representation. 
Mofasa tracks the QMOF training distribution well across all three dimensions, with a slight bias towards short $a$ \& $b$ lengths. The second row visualizes the lattice angles ($\alpha, \beta, \gamma$), where \modelname{} is again reasonable, but overemphasizes ${\sim}90^{\circ}$ angles. The third row examines symmetry and topology. We calculate spacegroups using \texttt{pymatgen} \cite{ongPythonMaterialsGenomics2013} at two different symmetry thresholds (symprec). The generated samples are $95\%$ and $84\%$ triclinic for $0.01{\AA{}}$ and $0.1{\AA{}}$ thresholds, which is substantially larger than the $54\%$ and $48\%$ in the data distribution. Surprisingly, geometry relaxation does not improve spacegroup symmetry matching, suggesting that the lack of diverse symmetries is a relatively fundamental problem, not simply a numerical precision issue.

The fourth row characterizes chemical complexity through the count of distinct nodes and linkers and the degree of catenation. In the distinct node/linker plots, the zero bin represents cases where the \texttt{MOFid} algorithm failed to identify a recognizable building block. The model overproduces failure cases and underproduces single node/linker systems. It also overproduces multivariate MOFs with $4+$ linkers, but otherwise matches the real data reasonably well. The final two rows present porosity metrics calculated using Zeo++ \citep{willems2012algorithms} with a $1.86$\angstrom{} Nitrogen probe. These four distributions show that the model generates MOFs with fairly realistic internal voids, but skews towards overly small LCD/PLD values, which in turn means a higher percentage of systems do not allow the Nitrogen probe to enter (see percentages in legends of final row plots).

\subsection{Dynamic behavior of sampled frameworks} \label{subsec:dynamic_mof}

\begin{figure}[h]
    \includegraphics[width=\linewidth]{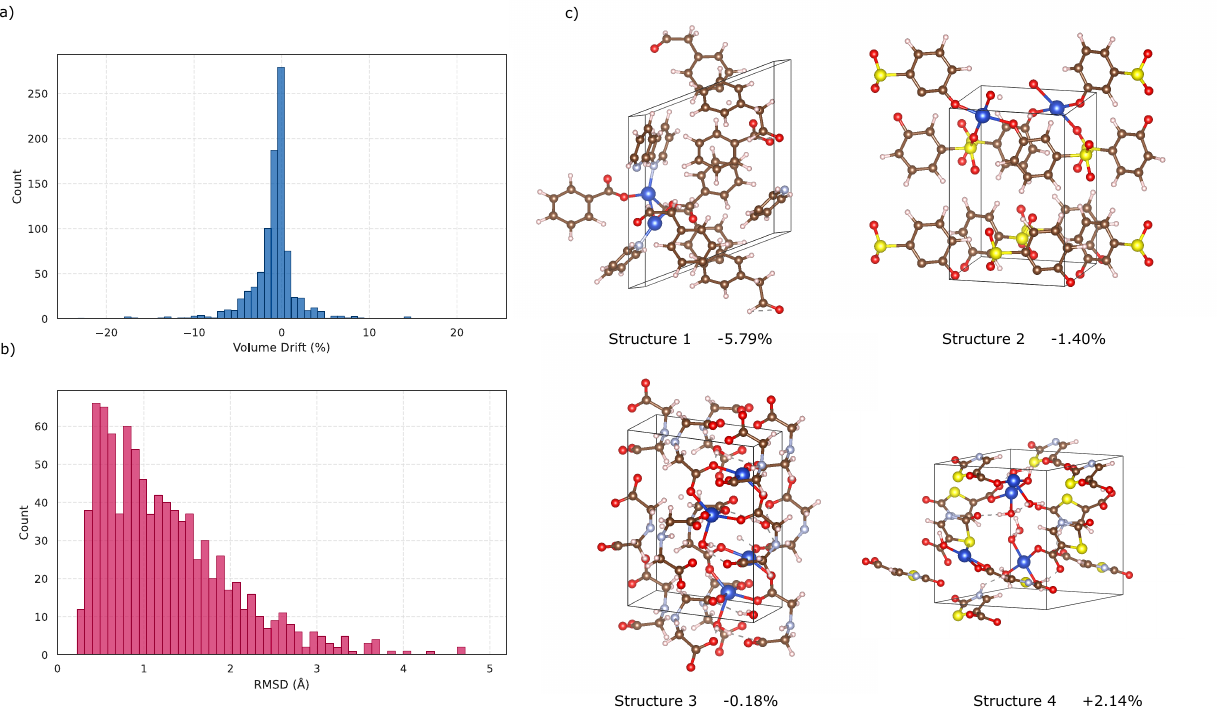}
    \caption{\textbf{Evaluation of dynamic stability.} Dynamic properties of 917 randomly selected samples from \dbname{}, calculated over a 50 ps MD trajectory. (a) Distribution of volume drift, calculated as the relative change in mean volume between the first and final 10 ps of the trajectory; (b) distribution of root mean square deviation (RMSD) between the initial and final frames; and (c) representative Cu-MOF trajectory endpoints with varying coordination geometries. The values below each structure indicate the volume drift percentage.}
    \label{fig:dynamic_mof}
\end{figure}

To investigate the dynamic behavior of generated MOFs, we used a short three-step MD pipeline (see Appendix \ref{sec:MD}) on 1,000 randomly selected samples from \dbname{}. These simulations indicate the structural flexibility of sampled structures above 0 K, helping us to understand dynamic behavior due to thermal effects that local geometry minimizations fail to identify. Our simulations used \texttt{orb-v3-con-inf-omat} \citep{neumannOrbFastScalable2024, rhodesOrbv3AtomisticSimulation2025} + D3 \citep{grimme2010consistent}, a combination that was shown to be highly accurate for MOFs in the zero-shot setting by \citet{krass2025mofsimbench}.

We report root mean square deviation (RMSD) of the initial and final frames of the NPT trajectory, as well as the volume drift between the first 10 ps and final 10 ps. Of the 1,000 samples, 3.6\% did not converge after 1,000 geometry optimization steps and 4.7\% exhibited rapid volume expansion, resulting in a failure of the MD run. The RMSD and volume drift histograms in \cref{fig:dynamic_mof}~(a-b) show that the majority of the samples do not exhibit a significant volume or atomic displacement, with 91\% drifting by less than 10\% in volume and 80.7\% having $< 2$\AA{} RMSD.

In \cref{fig:dynamic_mof}~(c), we highlight four representative copper-node systems, which demonstrate the wide range of coordination sites accessible to the generative model. The under-coordinated Structures 1 (T-shaped) and 2 (square planar) resemble less stable open metal sites, which resulted in the largest contraction during the NPT run. In contrast, Structure 3 adopted the thermodynamically preferred square pyramidal geometry, characteristic of the prototypical copper paddle-wheel,  and was the most rigid lattice. Finally, Structure 4 exhibited a distorted heteroleptic coordination environment, resulting in a volume expansion of 2.14\%.

Altogether, the high stability rate and chemically intuitive structures observed in the copper-node examples underscore the physical plausibility and coordination diversity of the samples.

\subsection{Conditional generation}
\begin{figure}
    \centering
    \includegraphics[width=0.9\linewidth]{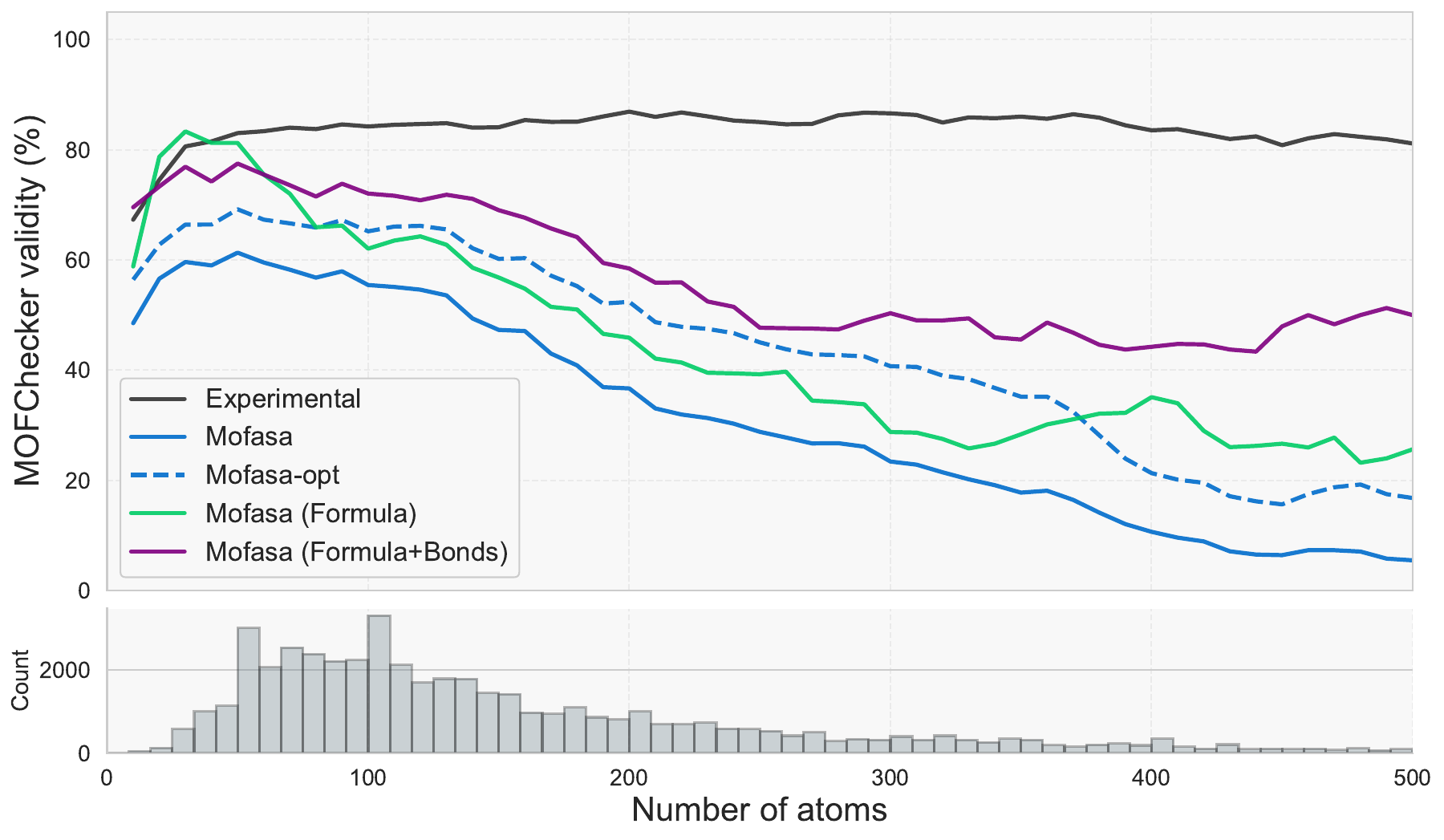}
    \caption{\textbf{MOFChecker validity vs system-size for different conditioning scenarios.} The more information that is conditioned on, the higher performance. The most informative conditioning (chemical formulas + bond-graphs) obtains $>40\%$ validity at 500 atoms.}
    \label{fig:system-scalability}
\end{figure}

All results presented so far pertain to unconditional samples. However, due to our multitask training setup (Section \ref{sec:key-model-choices}), \modelname{} can accept several types of conditioning, including i) per-node atom-types (chemical formula) and ii) a set of molecular graphs (formula + bonds) for each node/linker fragment. Both types of conditioning provide useful information to the model, which ought to enhance sample quality.

In Figure \ref{fig:system-scalability}, we confirm that conditioning increases \texttt{MOFChecker} validity across all system-sizes. We can distinguish three broad regimes: i) the $0-50$ atom range, where both types of conditioning boost performance significantly (+ $15$ - $20$ \%) ii) the $50-250$ atom range, where unconditional < formula < formula+bonds, with $\sim 10\%$ jumps between each iii) the $250-500$ atom range, which resembles the last case except with larger gaps of $15 - 20$ \%. At 500 atoms, unconditional \modelname{} decays to $5\%$, whilst formula + bonds remains steadily above $40\%$.

\section{Significance for generative modeling of materials}
\label{sec:Significance}

\modelname{} achieves high-fidelity generation---satisfying complex structural validity checks, exhibiting high degree of novelty, and maintaining high dynamic stability---all without relying on domain-specific compositional heuristics often used in MOF design.
This success is significant not just for MOF discovery, but for the broader field of generative models in materials science.

Recent successful models such as MOFDiff \cite{fuMOFDiffCoarsegrainedDiffusion2024}, MOFFUSION \cite{parkMultimodalConditionalDiffusion2025} and MOF-BFN \cite{jiaoMOFBFNMetalOrganicFrameworks2025} have relied on a modular approach, treating MOFs as compositions of rigid building blocks.
While this aligns with chemical intuition, it inherently constrains the model: architectures tailored to rigid building blocks are incompatible with non-modular domains.
This incompatibility isolates MOF generation from the broader landscape of generative models for materials and hinders the development of a foundation model for atomic systems.

To unlock cross-domain transfer learning, we must operate at the atomic granularity.
However, unified all-atom approaches, such as MatterGen \citep{zeniMatterGenGenerativeModel2025}, UniGenX \citep{zhangUniGenXUnifiedGeneration2025} and ADiT \citep{joshiAllatomDiffusionTransformers2025}, have, so far, not demonstrated scaling to large crystal systems like MOFs. They are often restricted to small crystal systems (${<}50$ atoms) or suffer severe underfitting when modeling complex porous crystals.
One might wonder if all-atom models suffer from a \emph{data bottleneck}---that they require massive datasets to learn how to generate valid structures.
Our results challenge this assumption by achieving state-of-the-art performance on systems up to 170 atoms using only ${\sim}14$k structures in QMOF (and up to 500 atoms on $49k$ experimental structures).

This efficiency validates the all-atom approach as a viable backbone for future foundation models: if \modelname{} can outperform domain-specific models, it is well-positioned to scale further. In the next section, we detail the implementation that enabled these results.

\section{An all-atom model for large crystal structure generation}
\label{section:methods}

\begin{figure}[t]
    \centering
    \includegraphics[width=1.0\linewidth]{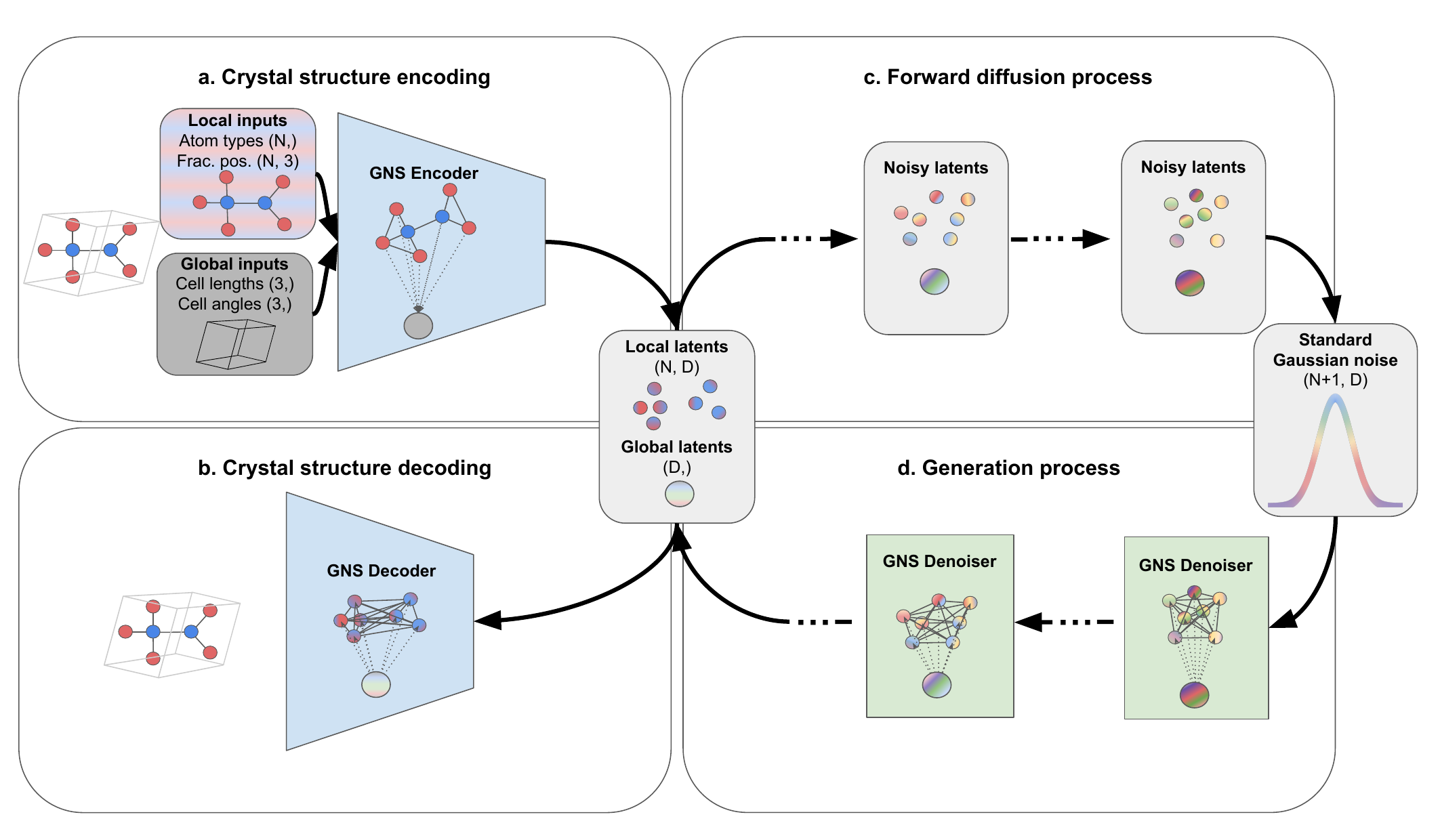}
    \caption{\textbf{Overview of \modelname{}.} (a) The encoder maps the crystal structure to a continuous latent representation. (b) The decoder reconstructs the structure from the latent representation. (c) The forward process corrupts the latents with noise until it is indistinguishable from standard Gaussian noise. (d) The denoising model learns to reverse this process to generate clean latents.}
    \label{fig:mofasa}
\end{figure}

\modelname{} follows the standard latent diffusion framework \citep{rombachHighResolutionImageSynthesis2022}, akin to recent generative approaches in materials science \cite{joshiAllatomDiffusionTransformers2025,luoUnifiedLosslessLatent2025}. 
As illustrated in \cref{fig:mofasa}, the model consists of three main components: encoder, decoder, and a denoising model. These components interact across three distinct stages:
\begin{itemize}
    \item \textbf{Autoencoding crystal structures (Panels~a, b):} The encoder maps an input crystal structure (atomic types, fractional positions, and lattice parameters) into a continuous latent representation. This representation consists of \emph{local latents}, encoding smoothed atomic neighborhoods, and \emph{global latents}, capturing system-wide properties. 
    \item \textbf{Latent diffusion (Panels~a, c, d):} We train a denoising diffusion model to generate valid latent representations. During training, clean latents from the encoder are gradually corrupted by noise until they are indistinguishable from a standard Gaussian distribution (Panel~c). 
    A denoising model is trained to predict the ``velocity'' required to reverse this corruption (Panel~d). 
    \item \textbf{Generation (Panels b, d):} To generate novel crystal structures, we sample pure Gaussian noise, iteratively denoise it using the trained model to obtain a clean latent representation (Panel~d), and then map it to the atomic domain using the decoder (Panel~b).
\end{itemize}

\subsection{Key Modeling Choices}
\label{sec:key-model-choices}

While the full methodology is detailed in \cref{sec:mofasa-detailed}, we highlight five important modeling decisions that enable high-fidelity generation.

\paragraph{Specialized processing for low- and high-level contexts.} 
The encoder, the decoder, and the diffusion model share a common backbone: a hierarchical graph network simulator (GNS) \citep{sanchez-gonzalezLearningSimulateComplex2020}.
Adapted from the Orb model \cite{rhodesOrbv3AtomisticSimulation2025}, we leverage an architecture validated for fast, high-accuracy interatomic potential modeling.
Importantly, the hierarchical message passing introduced here addresses the heterogeneity of crystal data, where local (atom-level) and global (system-level) features possess distinct properties necessitating specialized processing.
This separation also enables directional information flow: the encoder uses it to aggregate atomistic details into the global latents, whereas the decoder and denoiser use it to broadcast global context back to local representations.
This allows the model to simultaneously resolve fine-grained atomic details while maintaining precise long-range coherence.

\paragraph{Self-conditioning.} 
Although the learned latent space is continuous, it encodes mixed discrete (atom types) and continuous (fractional positions) structural information. 
We find that self-conditioning---often used to bridge the gap between discrete and continuous diffusion---is also essential in this context,
which we hypothesize is necessary for learning conditional dependencies between atom types (e.g.,\ common bonding pairs) \cite[][Appendix~A.3.2]{pynadathCANDIHybridDiscreteContinuous2025}.

\paragraph{Tweaked cosine schedule.}
We use a cosine-based schedule with a log-SNR shift. This modification allocates significantly more training capacity (approx. 78\% of diffusion timesteps) to the signal-dominated regime, which we found important for learning and generating fine-grained structural details.

\paragraph{Permutation symmetry breaking.}
Graph neural networks, such as our GNS backbone, are inherently permutation equivariant. 
However, the standard diffusion training objective regresses denoiser predictions against ordered targets, which introduces ambiguity: many permutations of the clean latents can result in the same noisy latents.
The network cannot distinguish between permuted versions of the graph, yet the loss penalizes them differently based on the ordered targets.
To resolve this ambiguity for \emph{de novo} generation, we explicitly break permutation symmetry during training by conditioning the model on a canonical node ordering derived from the crystal graph topology.
In contrast, we omit this ordering for conditional tasks (e.g.,\ conformer generation or structure inpainting), as the conditioning signal provides sufficient context to implicitly resolve the symmetry.

\paragraph{Multi-task conditional training.}
To enable broad applicability, we train a single model for multiple tasks. 
By stochastically providing conditioning information, \modelname{} learns to perform \emph{de novo} generation, structure inpainting (e.g.,\ generating linkers within a fixed node scaffold), and conditional generation based on chemical composition or bond topology.

\section{Related work}
\label{section:related}

\paragraph{Modular generation.} This approach exploits the compositional structure of MOFs by generating individual building blocks---nodes, linkers, or topology---before assembling them into a full structure \cite{park2024generative, fuMOFDiffCoarsegrainedDiffusion2024, parkMultimodalConditionalDiffusion2025}. 
GHP-MOFAssemble \cite{park2024generative} demonstrates this by generating linkers from molecular fragments using DiffLinker and combining them with pre-selected metal nodes into a fixed topology \cite{igashovEquivariant3DconditionalDiffusion2024}.
MOFDiff advances this via a coarse-grained (CG) diffusion process, where CG graph nodes, representing entire building blocks and their locations, are first generated via a diffusion model. Then, these CG nodes are replaced with atom-level details and assembled with a rigid assembly algorithm to determine the orientation of the otherwise fixed building block structure \cite{fuMOFDiffCoarsegrainedDiffusion2024}. However, MOFDiff is restricted to a pre-determined library of rigid building blocks found in the training distribution, and hence hence has a significantly limited generalization.
Addressing this limitation, MOFFlow-2 performs a similar two-step approach: first generating building block SMILES with an autoregressive model, then using a flow matching model infers the position and torsion angles for structural assembly \cite{kimFlexibleMOFGeneration2025}.
Finally, MOFFUSION departs from modeling the discrete building blocks, instead using latent diffusion to first generate a continuous signed distance function that defines the geometry of the MOF, and then decoding these shapes into atomic structures \cite{parkMultimodalConditionalDiffusion2025}.

\paragraph{All-atom \emph{de novo} generation.} 
A key requirement for crystal generation is the joint generation of atom types, positions, and lattice vectors. 
The seminal E(n)-equivariant diffusion model established the \emph{all-atom} framework for jointly generating atomic composition and geometry of \emph{molecules} directly in 3D space \cite{hoogeboomEquivariantDiffusionMolecule2022}. 
Simultaneously, CDVAE paved the way for a general-purpose all-atom approach to \emph{inorganic crystal} generation \cite{xieCrystalDiffusionVariational2022}.
Subsequently, MatterGen \cite{zeniMatterGenGenerativeModel2025} improved generation fidelity by scaling the diffusion principles to a diverse, curated dataset of crystal structures up to 20 atoms.
Simultaneously, UniMat \cite{yangScalableDiffusionMaterials2024a} introduced a periodic-table inspired representation for molecular and crystal structures and used a diffusion model to generate small crystals.
Flow-based model, such as FlowMM \cite{millerFlowMMGeneratingMaterials2024} and CrystalFlow \cite{luo2025crystalflow} have also been used for modeling crystal structures, offering increased modeling flexibility and faster sampling. 
However, despite these advancements, the all-atom class of models remained primarily used on small inorganic materials (typically fewer than 60 atoms per unit cell).

Addressing this gap, ADiT proposed a unified latent diffusion-based framework for generating both periodic materials and non-periodic molecules \cite{joshiAllatomDiffusionTransformers2025,luoUnifiedLosslessLatent2025}.
While promising for larger systems, ADiT focused on small systems (MP20, QM9), and its performance on complex porous crystals (QMOF) has been effectively limited (achieving $15.7\%$ \texttt{MOFChecker} validity on QMOF).
We adopt a similar diffusion-based framework as a conceptual foundation, re-engineering the model architecture, parameterization, and training objective. This yields state-of-the-art results on MOFs (achieving 52.9\% validity on QMOF and 52.2\% on a combination of experimental MOF DBs), while retaining a general-purpose capability applicable to diverse crystal modalities.

\begin{table}[t]
    \centering
    \caption{Comparison of generative models for Metal-Organic Frameworks and inorganic crystals.}
    \label{tab:related_work}
    \resizebox{\textwidth}{!}{%
    \begin{tabular}{@{}llll@{}}
        \toprule
        \textbf{Model} & \textbf{Approach} & \textbf{Resolution} & \textbf{Domain} \\ \midrule
        \multicolumn{4}{l}{\textit{Modular \& Assembly-based}} \\
        GHP-MOFAssemble \cite{park2024generative} & Diffusion (Linkers) + Rigid assembly & Building blocks & MOFs \\
        MOFDiff \cite{fuMOFDiffCoarsegrainedDiffusion2024} & Diffusion (CG) + Rigid assembly & Building blocks & MOFs \\
        MOFFLOW-2 \cite{kimFlexibleMOFGeneration2025} & Autoregressive (Linkers) + Flow-matching (Assembly) & Building blocks & MOFs \\
        MOFFUSION \cite{parkMultimodalConditionalDiffusion2025} & Latent diffusion (Mesh) & Implicit (Mesh) & MOFs \\ \midrule
        \multicolumn{4}{l}{\textit{LLM \& Agentic Approaches}} \\
        CrystaLLM \cite{antunes2024crystal} & Autoregressive & Text (CIF) & General \\
        MatterGPT \cite{chen2024mattergpt} & Autoregressive & Text (SLICES) & General \\
        MOFGPT \cite{badrinarayananMOFGPTGenerativeDesign2025} & Autoregressive & Text (\texttt{MOFid}) & MOFs \\
        Inizan et al. \cite{inizan2025system} & Agentic pipeline (LLM + Diffusion + Screening) & Hybrid (Text + SMILES + 3D) & MOFs \\ \midrule
        \multicolumn{4}{l}{\textit{Direct All-Atom Generation}} \\
        CDVAE \cite{xieCrystalDiffusionVariational2022} & VAE + Diffusion & All-atom & General \\
        MatterGen / UniMat \cite{zeniMatterGenGenerativeModel2025,yangScalableDiffusionMaterials2024a} & Diffusion & All-atom & General \\
        FlowMM / CrystalFlow \cite{millerFlowMMGeneratingMaterials2024, luo2025crystalflow} & Flow-matching & All-atom & General \\
        ADiT / UAE-3D \cite{joshiAllatomDiffusionTransformers2025,luoUnifiedLosslessLatent2025} & Latent diffusion & All-atom & General \\
        \textbf{\modelname{} (Ours)} & \textbf{Latent diffusion} & \textbf{All-atom} & \textbf{General} \\ \bottomrule
    \end{tabular}%
    }
\end{table}

\paragraph{LLM-based generation.} While diffusion-based approaches dominate 3D structure generation, autoregressive language models have also been applied to MOF crystal generation by treating structures as text sequences.
Models such as CrystaLLM \cite{antunes2024crystal} and MatterGPT \cite{chen2024mattergpt} are LLMs trained on crystallographic information files (CIF) and SLICES representations, respectively, learning the grammar of crystal structures. 
Similarly, MOFGPT uses an LLM to generate \texttt{MOFid} identifier sequences, incorporating a transformer-based property predictor and a reinforcement learning-based module for property-guided design \cite{badrinarayananMOFGPTGenerativeDesign2025}. 
Moving beyond direct sequence generation, \textcite{inizan2025system} recently proposed a promising agentic pipeline where an LLM orchestrates linker SMILES generation while a separate diffusion model handles 3D structure generation. The generated structures are then geometry-optimized and screened, to ensure validity and synthesizability.
By achieving a high-fidelity 3D generation directly, \modelname{} offers a way to significantly reduce the cost of the post-processing steps, potentially accelerating agentic materials discovery by orders of magnitude.

\section{Limitations and future work}

While \modelname{} achieves state-of-the-art performance in MOF generation, several limitations remain that present opportunities for future research.

An important constraint is the computational complexity of the architecture.
The decoder and denoiser models operate on fully-connected graphs because, unlike in the encoder, the atomic positions are undefined during the early stages of generation, preventing the use of nearest neighbor methods typically used to construct sparse graphs.
As a result, the memory requirement scales quadratically ($N^2$) with the number of atoms ($N$), making scaling beyond $N=500$ atoms per unit cell difficult.
Future work may investigate sparse graph approximations with long-range shortcut connections to maintain information flow while reducing the memory and computational overhead.

In addition, the GNS backbone architecture leaves room for optimization. We believe that explicit edge representations may be redundant. Moreover, the hierarchical message passing could be simplified to further reduce the number of computations.

Another key limitation is the requirement to specify the number of atoms ($N$) \emph{a priori}. 
In \emph{de novo} generation, this can be done by sampling $N$ from the empirical training distribution. 
However, this poses a challenge for conditional generation, where specific property constraints (e.g.,\ metal node composition) effectively alter the distribution of feasible atom counts. Currently, \modelname{} does not have a mechanism to sample this conditional distribution.

Finally, while \modelname{} is designed to be general-purpose, this study focused exclusively on Metal-Organic Frameworks. Investigating the model's performance and transfer learning between other material classes remains an important direction for future investigation.

\section{Conclusion}

We introduced \modelname{}, an all-atom generative model based on latent diffusion. 
We evaluated the model on Metal-Organic Framework (MOF) generation---a domain of structurally complex porous materials---valued for carbon capture, gas separation, catalysis, and hydrogen storage.
\modelname{} achieves unprecedented generation fidelity, demonstrating high validity and dynamic stability.
Notably, the model shows strong generalization, rediscovering metal nodes and topologies absent from the training distribution but observed in other experimental datasets.

\modelname{} is the first model to successfully scale the all-atom approach to large crystal structures. We demonstrated the generation of complex unit cells containing up to 500 atoms, scaling well beyond prior inorganic crystal models (typically limited to $<60$ atoms). 
Unlike tailored approaches that rely on MOF-specific decompositions, our method provides a more flexible framework, enabling the discovery of more novel chemistries.

Ultimately, these results extend beyond the MOF domain.
By eliminating the need for domain-specific heuristics while maintaining high structural fidelity at scale, \modelname{} validates the all-atom diffusion approach as a robust path toward future foundation models capable of transfer learning across the full range of materials.

\newpage
\printbibliography

\begin{appendices}
  \titleformat{\section}
    {\normalfont\Large\bfseries}
    {Appendix \thesection:}
    {1em}
    {}
    
\section{MofasaDB}
\label{sec:mofasadb}

To help the scientific community build on top of this work, we provide a dataset of 201,926 raw and Orb-optimized structures generated by \modelname{-QMOF} (trained on the QMOF (${<}170$ atoms) subset). 
To enable property-based screening, all samples include the pre-computed properties detailed in \cref{sec:property-glossary}. For each sample, we additionally include latent embeddings from the \texttt{orb-v3-direct-20-omat} MLIP \cite{rhodesOrbv3AtomisticSimulation2025}. The data can be accessed at at \url{https://huggingface.co/datasets/Orbital-Materials/MofasaDB} and is released under the CC-BY 4.0 license.

Additionally, we also make available an interactive web interface for exploring and screening the unoptimized database; available at \url{https://mofux.ai/}. The exploration is supported by the included latents, allowing interactive search in the \texttt{Orb} latent space, which clusters similar MOFs as demonstrated in \ref{sec:latent-embeddings}. 

\section{Training datasets \& pre-processing}
\label{sec: training data}

Here, we detail the datasets used to train \modelname{} and the relevant baselines.
All crystal structures used to train \modelname{} were standardized to their primitive unit cells using \texttt{pymatgen} \cite{ongPythonMaterialsGenomics2013}. Additionally, we applied the \texttt{metal-oxo} algorithm from the \texttt{MOFid} package \cite{buciorIdentificationSchemesMetal2019} to decompose the MOF structures.
This generated atom-level labels (metal nodes, bridges, organic linkers, and solvents) which were used for conditional training and downstream analysis.

\paragraph{ADiT + QMOF.}
The ADiT baseline \cite{joshiAllatomDiffusionTransformers2025} was trained on the Quantum MOF (QMOF) database \cite{rosenMachineLearningQuantumchemical2021}, which contains $20$k experimentally synthesized MOFs with structures relaxed via DFT. The authors filtered the dataset to include only structures with fewer than 150 atoms in the unit cell (corresponding to the 81st percentile of the full distribution). This resulted in a final training set of approximately $14$k samples after a training/validation/test split.

\paragraph{\modelname{} + QMOF.}
To compare against the ADiT baseline, we trained \modelname{} on QMOF systems with up to 170 atoms in the unit cell (corresponding to 85th percentile). The data was randomly split 80/10/10\% into training, validation, and test sets, resulting in a final training dataset of approximately $14$k structures.

\paragraph{MOFFlow-2 + BW.}
The MOFFlow-2 baseline \cite{kimFlexibleMOFGeneration2025} was trained on the Boyd-Woo (BW) database \cite{boydComputationalDevelopmentNanoporous2017,boydDatadrivenDesignMetal2019}, consisting of $358$k total hypothetical MOFs constructed using the ToBasCCo assembly method \cite{boydGeneralizedMethodConstructing2016a}.
The authors used \texttt{MOFid} to decompose MOFs into the building blocks, discarded any structures with more than 20 building blocks, and filtered out all systems that failed \texttt{MOFChecker} criteria. This resulted in a training dataset of $157$k samples with 100\% \texttt{MOFChecker} validity.

\paragraph{MOF-BFN + BW.}
The MOF-BFN baseline \cite{jiaoMOFBFNMetalOrganicFrameworks2025} also used the BW database. In contrast to MOFFlow-2 they only filtered out structures with more than 200 building blocks. The authors did not report the exact final training set size.

\paragraph{\modelname{} + BW.}
For comparison with MOFFlow-2 \cite{kimFlexibleMOFGeneration2025} and MOF-BFN \cite{jiaoMOFBFNMetalOrganicFrameworks2025} we use \modelname{} on ARC-MOF DB0 split \cite{burnerARCMOFDiverseDatabase2023}, which consists of $263$k systems after additional ARC-MOF structure checks, including metal oxidation states, atom overlaps, unrealistically small unit cells, and over-coordination. 
We further filter out systems that have more than 500 atoms in the unit cell.
After a 95/4/1\% training/validation/test split, we trained \modelname{} on $238$k systems.

\paragraph{\modelname{} + Experimental.}
We define the ``Experimental'' dataset as a combination of several available experimental sources: QMOF, the computatioSn-ready split of CoRE-MOF-2024 \cite{zhaoCoREMOFDB2025}, and the DB12 and DB14 subsets from ARC-MOF \cite{burnerARCMOFDiverseDatabase2023}. This combinations results in a total of $49$k systems with an average \texttt{MOFChecker} validity of 84.2\%. When training \modelname{} we only use systems up to 500 atoms from the Experimental dataset.

\section{Building \modelname{}}
\label{sec:mofasa-detailed}

We adopt the latent diffusion framework \cite{rombachHighResolutionImageSynthesis2022} involving an autoencoder and a denoising diffusion model. 
The autoencoder learns a continuous latent representation from a crystal structure's mixed categorical and continuous features, while the diffusion model learns to generate these latent representations. The autoencoder and diffusion model components share a common graph neural network (GNN) backbone, which is an adaptation of an established architecture previously validated in machine-learned interatomic potential models \cite{neumannOrbFastScalable2024,rhodesOrbv3AtomisticSimulation2025} for fast and accurate molecular simulations.

\subsection{Hierarchical GNS backbone}
\label{sec:gns}

The shared backbone architecture for both autoencoder and diffusion models is an extended version of the Graph Network Simulator (GNS) \cite{sanchez-gonzalezLearningSimulateComplex2020}. 
This architecture incorporates two key modifications: (1) the addition of a graph attention mechanism \cite{velickovicGraphAttentionNetworks2018}, and (2) a hierarchical message passing scheme. 
This hierarchical scheme is motivated by the heterogeneity of atomistic structural data, where structural features at the \emph{local} and \emph{global} levels have fundamentally distinct properties, hence necessitating specialized processing. 
We enable this by defining the graph components as distinct local and global sets, where local components may represent individual atoms (or atomic neighborhoods) and global components capture broader system properties like lattice parameters.

We denote the sets of local nodes, local edges, global nodes, and global edges as $\nodes_{\text{L}} = \{\node^\text{L}_{i}\}$, $\edges_{\text{L}} = \{\edge^\text{L}_{j,k}\}$, $\nodes_{\text{G}} = \{\node^\text{G}_{l}\}$, and $\edges_{\text{G}} = \{\edge^\text{G}_{m,n}\}$, respectively. 
All edges are directional. While local edges are restricted to connecting local nodes, global edges are allowed to cross-communicate between the global and local contexts, handled by a dedicated neural network distinct from the local message passing networks. Precise construction of these sets will be introduced in subsequent sections.

\begin{algorithm}[tbp]
    \caption{Hierarchical GNS processing}
    \label{alg:gns-processing}
    \begin{algorithmic}[1]
    \Require
    \Statex \textbf{Features:} Local nodes $\nodes_{\text{L}}^{0}$ and edges $\edges_{\text{L}}^{0}$, and global nodes $\nodes_{\text{G}}^{0}$ and edges $\edges_{\text{G}}^{0}$.
    \Statex \textbf{Embedders:} Local embedding MLP, $\textsc{Embed}_{\text{L}}(\cdot)$, and global embedding MLP, $\textsc{Embed}_{\text{G}}(\cdot)$.
    \Statex \textbf{Message passing networks:} Local edge and node networks, $\phi_{\text{E,L}}^{t}(\cdot)$ and $\phi_{\text{N,L}}^{t}(\cdot)$, and global edge and node networks, $\phi_{\text{E,G}}^{t}(\cdot)$ and $\phi_{\text{N,G}}^{t}(\cdot)$ for $t \in \{1, \ldots, T\}$.
    \Statex \textbf{Edge-attention functions:} $\psi_{1}^{t}(\cdot)$ and $\psi_{2}^{t}(\cdot)$ for $t \in \{1, \ldots, T\}$.
    \Statex \textbf{Read-out layers:} Local read-out MLP, $\textsc{Read}_{\text{L}}(\cdot)$, and global read-out MLP, $\textsc{Read}_{\text{G}}(\cdot)$.
    \Statex
    \Function{ComputeDeltas}{$\nodes, \edges$; $\enn{}$, $\nnn{}$, $\psi_{1}$, $\psi_{2}$}
        \For{\textbf{each} $\edge_{i,j} \in \edges, \node_{i} \in \nodes, \node_{j} \in \nodes$:} \Comment{Compute edge residuals for each edge}
            \State $\Delta \edge_{i,j} \leftarrow \enn{}([\edge_{i,j}, \node_{i}, \node_{j}])$ 
        \EndFor
        \For{\textbf{each} $\node_{i} \in \nodes$:} \Comment{Compute node residuals for each node}
            \State $\m_{i}^{1} \leftarrow \sum_{j \in \mathcal{R}(i)} \psi_{1}(\edge_{i,j}) \Delta \edge_{i,j}; \qquad \m_{i}^{2} \leftarrow \sum_{j \in \mathcal{S}(i)} \psi_{2}(\edge_{j,i}) \Delta \edge_{j, i}$
            \State $\Delta \node_{i} \leftarrow \nnn{}([\node_{i}, \m_{i}^{1}, \m_{i}^{2}])$
        \EndFor
        \State \textbf{return} $\{\Delta \node_i\}_{i}$, $\{\Delta \edge_{i,j}\}_{i,j}$
    \EndFunction
    \Statex
    \Statex \color{gray}\text{\# 1. Embedding Stage} \color{black}
    \State $\nodes_{\text{L}}^1, \edges_{\text{L}}^1 \leftarrow \textsc{Embed}_{\text{L}}(\nodes_{\text{L}}^0, \edges_{\text{L}}^0)$; \qquad $\nodes_{\text{G}}^1, \edges_{\text{G}}^1 \leftarrow \textsc{Embed}_{\text{G}}(\nodes_{\text{G}}^0, \edges_{\text{G}}^0)$

    \Statex
    \Statex \color{gray}\text{\# 2. Message Passing Loop} \color{black}
    \For{$t = 1$ \textbf{to} $T$}
        \Statex \hspace{\algorithmicindent}\color{gray}\text{\# Local Step: Update local components}\color{black}
        \State $\Delta \nodes_{\text{L}}^{t}, \Delta \edges_{\text{L}}^t \leftarrow \textsc{ComputeDeltas}(\nodes_{\text{L}}^t, \edges_{\text{L}}^t; \lenn{t}, \lnnn{t}, \psi_{1,\text{L}}^t, \psi_{2,\text{L}}^t)$
        \State $\nodes_{\text{L}}^{t+1} \leftarrow \nodes_{\text{L}}^{t} + \Delta \nodes_{\text{L}}^{t}; \qquad \edges_{\text{L}}^{t+1} \leftarrow \edges_{\text{L}}^{t} + \Delta \edges_{\text{L}}^{t}$
        
        \Statex
        \Statex \hspace{\algorithmicindent}\color{gray}\text{\# Global Step: Update global components, incorporating local node info}\color{black}
        \State $\nodes_{\text{Joint}}^{t} \leftarrow \nodes_{\text{G}}^{t} \cup \nodes_{\text{L}}^{t+1}$
        \State $\Delta \nodes_{\text{Joint}}^{t}, \Delta \edges_{\text{G}}^t \leftarrow \textsc{ComputeDeltas}(\nodes_{\text{Joint}}^{t}, \edges_{\text{G}}^t; \genn{t}, \gnnn{t}, \psi_{1,G}^t, \psi_{2,G}^t)$
        \State $\nodes_{\text{G}}^{t+1} \leftarrow \nodes_{G}^{t} + \Delta \nodes_{\text{Joint} \mid \text{G}}^{t}; \qquad \edges_{\text{G}}^{t+1} \leftarrow \edges_{\text{G}}^{t} + \Delta \edges_{\text{G}}^{t}$

        \Statex
        \Statex \hspace{\algorithmicindent}\color{gray}\text{\# Broadcast global information to connected local nodes}\color{black}
        \State $\nodes_{\text{L}}^{t+1} \leftarrow \textsc{ScatterAdd}(\nodes_{\text{L}}^{t+1}, \Delta \nodes_{\text{Joint} \mid \text{L}}^{t})$
    \EndFor

    \Statex
    \Statex \color{gray}\text{\# 3. Read-out Stage} \color{black}
    \State \Return $\textsc{Read}(\nodes_{\text{L}}^{T+1}), \textsc{Read}(\nodes_{\text{G}}^{T+1})$
    \end{algorithmic}
\end{algorithm}

The processing of the GNS is outlined in \cref{alg:gns-processing}, with the key modification being the message passing stage. 
Each message passing iteration consists of two sequential steps: a local step, where information is exchanged exclusively between local nodes via local edges; and a global step, which processes a joint graph of global and local nodes. This global step aggregates atom-level information to update the global context and, simultaneously, broadcasts system-level context to the local nodes connected via the directed global edges.
The message computation also involves an edge-attention mechanism, adopted directly from Orb \citep[][Section~2.2]{neumannOrbFastScalable2024}.

\subsection{Learning to represent crystal systems}
\label{sec:autoencoder}

A crystal structure is defined in terms of its periodically repeating unit cell containing $N$ atoms. To ensure a unique representation across infinitely-many arbitrary unit cell choices, we standardize the structures using primitive cell reduction and Niggli reduction \cite{grosse-kunstleveNumericallyStableAlgorithms2004}. 
We represent this standardized structure with the tuple $\system = (\atoms, \fpositions, \lattice)$, where $\atoms = \{\atom_i\}_{i=1}^{N} \in \atomdomain^{N}$ denotes the atomic types, $\fpositions = \{\fposition_i\}_{i=1}^{N} \in \unitdomain^{N \times 3}$ represents the fractional positions relative to the unit cell, and $\lattice = (a, b, c, \alpha, \beta, \gamma) \in \posdomain^{3} \times [\frac{\pi}{3}, \frac{2\pi}{3}]^{3}$ is the rotationally-invariant lattice representation consisting of three length parameters and three angles describing the basis vectors.

While the above representation is rotationally invariant, the fractional positions depend on the arbitrary choice of the unit cell origin. 
To ensure the model learns robust representations, we use data augmentation.
For every training sample, we shift all fractional positions by a random vector $\bm{u} \sim \mathcal{U}\unitdomain^{3}$, obtaining the augmented sample $\system' = (\atoms, (\fpositions + \bm{u}) \bmod 1, \lattice)$, where the modulo operator corresponds to the periodic boundary condition and ensures that the fractional positions remain in $\unitdomain^{3}$.
For conciseness in this paper we will use $\system$ and $\system'$ interchangeably, since they represent the same system.

Learning to directly generate mixed categorical and continuous data, such as crystal structures, presents significant challenges \citep[e.g.,][]{nazabalHandlingIncompleteHeterogeneous2020,hoogeboomEquivariantDiffusionMolecule2022,kotelnikovTabDDPMModellingTabular2023,xuGeometricLatentDiffusion2023}. 
To mitigate these issues, we train an autoencoder that maps the crystal structure $\system$ to a continuous latent representation $\latents$ using an encoder $\enc(\cdot)$, and reconstructs it via a decoder $\dec(\cdot)$. 

\subsection{Mapping crystal systems to latent representations}

The encoder $\enc(\cdot)$ maps a crystal structure $\system$ to a continuous latent representation $\latents$ using the GNS architecture defined in \cref{sec:gns}. We adapt the implementation and input featurization from Orb \cite{neumannOrbFastScalable2024,rhodesOrbv3AtomisticSimulation2025}, a state-of-the-art MLIP, leveraging its ability to produce robust representations for modeling potential energy and atomic forces.

\paragraph{Feature construction.} We construct the input local node features $\nodes_{\text{L}}$ by concatenating the one-hot representations of the atomic types $\atoms$ with sinusoidal embeddings of the fractional coordinates $\fpositions$. 
The input local edges $\edges_{\text{L}}$ are then constructed using a unit cell-aware nearest neighbor scheme; the resulting edge displacement vectors are encoded using Bessel basis functions. 
To incorporate lattice information in the local context, we concatenate the lattice parameters, encoded via radial basis functions, to both the local node and edge feature vectors.
Finally, we initialize a single global node $\nodes_{\text{G}}$ using the encoded lattice parameters and create directed global edges pointing from every local node to this global node. 

\paragraph{Encoder processing.} The GNS message passing maps these inputs to the latent representation $\latents$. The local message passing layers, which incorporate distance-smoothed attention \cite{neumannOrbFastScalable2024}, learn a smooth, localized representation of atomic structure, denoted $\llatents \in \realdomain^{N \times D}$. 
On the other hand, the global message passing steps aggregate system-level context from both local and global nodes into a global latent representation $\glatents \in \realdomain^{D}$. Together, these define the continuous latent representation as $\latents = (\llatents, \glatents)$, which we learn to generate using a diffusion model in \cref{sec:latent-diffusion}.

\subsection{Regularizing the representation space}

To ensure the latent space is structured and informative for the diffusion model, we regularize representations with a bottleneck layer before the decoder. 
We set the latent dimensionality to $D = 4$.
We use separate bottleneck layers for the two components: for the local latents $\llatents$ we use residual vector quantization \cite{zeghidourSoundStreamEndtoEndNeural2021} with the rotation trick \cite{fiftyRestructuringVectorQuantization2025}, whereas for the global latent vector $\glatents$, we apply a Gaussian KL-divergence bottleneck. 
The regularized features after the bottleneck are denoted as $\widetilde \latents = \bottleneck(\latents)$.
Empirically, we observed that the model performance is robust to the specific choice of bottleneck configuration.

\subsection{Mapping latent representations to crystal systems}

Similar to the encoder, the decoder model $\dec(\cdot)$ uses the GNS architecture to reconstruct a crystal structure $\hat \system$ from the latent representation $\widetilde \latents$.

\paragraph{Feature construction.} We initialize the input node features using the regularized latents $\widetilde \latents$ obtained from the bottleneck layer. 
In contrast to the sparse nearest-neighbor graph used in the encoder, the decoder constructs a fully-connected graph for the local context: local edges are created from each local node to every other node, initialized with the concatenated latent vectors of the connected pair. 
Additionally, the directed global edges are initialized from the single global node to every local node.

\paragraph{Decoder processing.} Decoder message passing uses the edge-attention mechanism \cite{neumannOrbFastScalable2024}, omitting the distance-based smoothing used in the encoder, as spatial positions are not yet defined at this stage.
The local message passing steps evolve the local features to recover atom-level details, while the global steps broadcast system-level context to the local nodes.
Finally, dedicated read-out heads produce the reconstruction parameters: the fractional coordinates $\hat \fpositions$ and atom type probabilities $ \hat{\bm{P}}_{\atoms}$ are predicted from local nodes (where the discrete atom type is given by $\hat \atoms = \{\argmax_j \hat{P}^{j}_{i}\}_i$), and the lattice parameters $\hat \lattice$ are predicted from the global node.

\subsection{Training the autoencoder}

We train the autoencoder to reconstruct the input crystal system $\system = (\atoms, \fpositions, \lattice)$ by minimizing a weighted sum of reconstruction losses and regularization losses.
Given the decoder outputs $(\hat{\bm{P}}_{\atoms}, \hat \fpositions, \hat \lattice) = \dec(\widetilde \latents)$ for latents $\widetilde \latents = \bottleneck(\latents)$ with $\latents = \enc(\system)$, we define the objectives as follows.

\paragraph{Reconstruction losses.} We use the average cross-entropy loss to predict the discrete atom types $\hat \atoms$:
\begin{align}
    \loss_{\atoms}(\atoms, \hat{\bm{P}}_{\atoms}) = \frac{1}{N} \sum_{i=1}^N \textsc{CrossEntropy}(\atom_i, \hat{\mathbf{P}}_i).
\end{align}
For fractional positions $\hat \fpositions$ we use mean squared error (MSE):
\begin{align}
    \loss_{\fpositions}(\fpositions, \hat \fpositions) = \frac{1}{N} \sum_{i=1}^N \norm{\fposition_i - \hat \fposition_i}_2^2.
\end{align}
To predict the lattice parameters $\hat \lattice = (\hat a, \hat b, \hat c, \hat \alpha, \hat \beta, \hat \gamma)$ we also use the MSE after normalization. We normalize the lattice vector lengths by the cube root of the atom count, $N^{1/3}$, following \textcite[Appendix~B.1]{xieCrystalDiffusionVariational2022}, and apply a logarithmic transformation, before computing the loss:
\begin{align}
    \loss_{\lattice}(\lattice, \hat \lattice) = \frac{1}{3} \norm{ \log\left(\frac{(a, b, c)}{N^{1/3}}\right) - \log\left(\frac{(\hat a, \hat b, \hat c)}{N^{1/3}}\right) }_2^2 + \frac{1}{3} \norm{(\alpha, \beta, \gamma) - (\hat \alpha, \hat \beta, \hat \gamma)}_2^2.
\end{align}

\paragraph{Regularization.} As discussed, we use different regularization methods for the local and global latent spaces. 
The local latents $\llatents$ use a residual vector quantization (VQ) bottleneck. We implement this using the \texttt{vector\_quantize\_pytorch} Python package\footnote{\url{https://github.com/lucidrains/vector-quantize-pytorch/}}, where the codebook $\bm{C}$ is updated using exponential moving average, and the encoder outputs are regularized to stay close to the codebook vectors via the standard commitment loss \cite{vandenoordNeuralDiscreteRepresentation2017}, ensuring the latent space does not grow during training:
\begin{align}
    \loss_{\text{L}}^{\text{Reg}}(\llatents) = \frac{1}{N} \sum_{i=1}^N \textsc{Commitment}(\latent_{\text{L}, i}, \bm{C}).
\end{align}
The global latents $\glatents$ are regularized via the KL-divergence:
\begin{align}
    \loss_{\text{G}}^{\text{Reg}}(\glatents) = \frac{1}{N} \KLD{\mathcal{N}(\glatents^{\bm{\mu}}, \text{diag}(\glatents^{\bm{\sigma}}))}{\mathcal{N}(\bm{0}, \bm{1})},
\end{align}
where the global latents correspond to the variational posterior parameters $(\glatents^{\bm{\mu}}, \glatents^{\bm{\sigma}}) = \glatents$, and $\mathcal{N}(\bm{0}, \bm{1})$ denotes the standard Gaussian prior.

\paragraph{Total objective.} The full training objective is the weighted sum of these components:
\begin{align}
    \loss_{\text{AE}}(\system, \hat \system, \latents) = 
    \begin{pmatrix}
        \loss_{\atoms}(\atoms, \hat{\atoms}) \\
        \loss_{\fpositions}(\fpositions, \hat{\fpositions}) \\
        \loss_{\lattice}(\lattice, \hat{\lattice}) \\
        \loss_{\text{L}}^{\text{Reg}}(\llatents) \\
        \loss_{\text{G}}^{\text{Reg}}(\glatents)
    \end{pmatrix}^{\top}
    \cdot
    \begin{pmatrix}
        1 \\
        300 \\
        1 \\
        1 \\
        10^{-4}
    \end{pmatrix}.
\end{align}

\subsection{Learning to generate latent representations of crystal structures}
\label{sec:latent-diffusion}

We use latent diffusion \cite{rombachHighResolutionImageSynthesis2022} to model the distribution of crystal structure latent representations. 
Once trained, the model generates samples of the latents $\latents$, which are mapped back to the atom domain using the decoder: $\hat \system = \dec(\bottleneck(\latents))$.

We formulate the generative process using a denoising diffusion probabilistic model (DDPM) \cite{hoDenoisingDiffusionProbabilistic2020} operating on the continuous latent space $\latents \in \realdomain^{(N+1) \times D}$.
By modeling crystal structures in this continuous space, we avoid the challenges posed with directly generating mixed categorical and continuous data in the atom domain.

Since the dimensionality of the latent space depends on the system size (comprising $N$ local nodes and 1 global node), the diffusion process is explicitly conditional on the number of atoms $N$. 
During inference, $N$ is sampled from the empirical distribution of the training data. 
The diffusion framework is defined by two processes: a fixed forward process that gradually corrupts the data structure by adding noise, and a learnable reverse process that learns to generate data from pure noise.

\paragraph{Data preparation.}
The training targets are generated on-the-fly by passing the crystal systems $\system$ through the frozen encoder $\enc(\cdot)$. 
To help the model learn a translationally invariant distribution, we apply the random translation augmentation described in \cref{sec:autoencoder} to the input structures before encoding. 
Moreover, to stabilize diffusion model training, we standardize the latent representations to have a zero mean and unit variance using running statistics. For conciseness, we re-use $\latents$ to denote the \emph{standardized} latents in the remainder of this section.

\paragraph{Forward process.} 
We define the forward process that transforms the clean latent representation $\latents_0$ (obtained from the encoder) into a standard Gaussian distribution over a sequence of timesteps $t \in \{0, \ldots, T\}$ (we set $T=100$k during training, and $T=4$k during sampling). 
The noisy sample at timestep $t$, denoted $\latents_t$, is sampled as:
\begin{align}
    \latents_t = \sqrt{\baralpha_t} \latents_0 + \sqrt{1-\baralpha_t} \eps, \qquad \text{where } \eps \sim \mathcal{N}({\bm{0}, \bm{1}}),
\end{align}
where $\eps$ has the same dimensionality as the latents. The noise schedule $\bar \alpha_t$ follows a cosine-based schedule \cite{hoDenoisingDiffusionProbabilistic2020}:
\begin{align}
    \baralpha_t = \text{sigmoid}\left(\log \text{SNR}(t)\right), \quad \text{with } \quad \log \text{SNR}(t) = -\log \tan^2 \left(\frac{t\pi}{2T}\right) + s
\end{align}
where we introduce a shift hyperparameter $s=2$ in the log signal-to-noise ratio (SNR) domain, ensuring that the diffusion model spends more time in the regime dominated by signal than noise (with $s=2$ this corresponds to 78\% of the diffusion timesteps).
As $t \rightarrow T$, the SNR ratio approaches zero, ensuring that the latents $\latents_T$ are indistinguishable from pure Gaussian noise.

\paragraph{Reverse process.}
The reverse diffusion process generates latents $\hat \latents_0$ from pure noise $\hat \latents_T$. 
Here, a denoising model is trained to reverse the corruption step-by-step. 
We use the $v$-prediction (``velocity'') parameterization \cite{salimansProgressiveDistillationFast2022}. The denoising model predicts $\vb_t$, which represents the velocity in the latent space, defined as a linear combination of the clean latents and the noise:
\begin{align}
    \vb_t = \sqrt{\baralpha_t} \eps - \sqrt{1 - \baralpha_t} \latents_0
\end{align}
This parametrization improves training stability and convergence, particularly because the target remains well-defined at $t=T$ (where $\text{SNR}=0$), where the standard $\eps$-prediction objective is generally unstable due to $\text{SNR}=0$ \cite{salimansProgressiveDistillationFast2022}.

\paragraph{Self-conditioning.}
We implement self-conditioning \cite{chenAnalogBitsGenerating2023} to improve training convergence and sample quality.
During training, with probability of 0.5, we condition the denoising model on a noisier sample from a later stage in the diffusion process. 
We sample a time offset $\delta t \sim \mathcal{U}[1, \ldots, \lfloor T*0.002\rfloor]$ and define an auxiliary timestep $t' = \min(t + \delta t, T)$.
The model is then conditioned as $\vb_{\thetab}(\latents_t, t, \latents_{t'}, t')$, where 
\begin{align}
    \latents_{t'} = \sqrt{\frac{\baralpha_{t'}}{\baralpha_t}}\latents_t + \sqrt{1 - \frac{\baralpha_{t'}}{\baralpha_t}}\eps', \quad \text{with } \quad \eps' \sim \mathcal{N}(\bm{0}, \bm{1}).
\end{align}
This random sampling of the auxiliary timestep $t'$ is different from the original implementation and allows the model to generalize to varying diffusion schedules and timestep granularity during sampling. 
Empirically, we found self-conditioning to be essential for model performance.
We hypothesize that this necessity arises from the nature of the latent space: although the representations are continuous, they encode inherently discrete structural information. 
As recently suggested by \textcite{pynadathCANDIHybridDiscreteContinuous2025}, self-conditioning enables learning conditional dependencies, such as, atom bonding pairs, proving critical in this mixed latent space.

\paragraph{Training objective.} 
The training objective for the denoising model $\vb_{\thetab}(\latents_t, t, \latents_{t'}, t')$ is the expected MSE:
\begin{align}
    \loss_{\text{Diff}} = \mathbb{E}_{t, \latents_0, \eps, t', \latents_{t'}, \eps'} \norm{\vb_t - \vb_{\thetab}(\latents_t, t, \latents_{t'}, t')}_2^2, \label{eq:diffusion-loss}
\end{align}
where $t \sim \mathcal{U}[0, \ldots, T]$, $\latents_0$ are systems $\system$ from the empirical distribution encoded using the fixed encoder, $\eps$ is noise from standard Gaussian distribution that has the same shape as $\latent_0$, and the self-conditioning information $t'$, $\latents_{t'}$ and $\eps'$ is constructed as described above.

\paragraph{Denoising GNS model.}
We use the same GNS backbone described in \cref{sec:gns} for the denoising model $\vb_{\thetab}(\latents_t, t, \latents_)$. 
We concatenate the $\latents$ with a positional embedding of the time index $t$ and pass them into the GNS, which predicts the velocity $\vb_{t}$.

\paragraph{Permutation symmetry breaking for \emph{de novo} generation.}
The GNS backbone is inherently permutation-equivariant; that is, permuting the input node ordering results in an equivalent permutation of the output features. 
The standard diffusion objective in \cref{eq:diffusion-loss} regresses these order-dependent model predictions against ordered targets. 
However, in high noise regimes, the correspondence between the noisy latents and the clean data becomes ambiguous, as multiple node orderings could have generated the same noisy latents.
This creates a difficult optimization task since an order-agnostic denoising model cannot effectively match the fixed target ordering.
While recent works propose matching-based objectives to ensure a permutation-invariant loss \cite{tongRaoBlackwellGradientEstimators2025}, we choose a simpler approach: we explicitly break the symmetry within the denoising model itself.
We achieve this by concatenating embeddings of the node order index to the node latent features before passing them to the GNS. 
Empirically, we observe that this order-conditioning is important for performance in the \emph{de novo} setting. In contrast, it is less significant for conditional generation, where the conditioning signal often provides sufficient structural context to implicitly resolve the symmetry.

To maximize the efficacy of this symmetry breaking, it is important that the node order is consistent across data.
We achieve this by pre-processing the data to enforce a canonical ordering derived from the crystal graph's topology.
Specifically, we construct the bond graph of the system and determine the node order via the BLISS graph coloring algorithm \cite{junttilaEngineeringEfficientCanonical2007}, using atom types as the initial node colors to distinguish chemically distinct atoms.
For MOFs, we make a further adjustment: we use the \texttt{MOFid} algorithm first to decompose the structure into its constituent building blocks (linkers, solvents, metal nodes and bridges) and apply the graph canonicalization algorithm within each component independently. 

During \emph{de novo} sampling, we condition the denoising model on an index sequence $[1, \ldots, N]$. Since this sequence relies only on the system size $N$ and provides no prior information regarding the crystal's chemistry or topology, the generative process is unconditional.

\paragraph{Conditional generation.}
A key advantage of the direct mapping between the latents $\latents \in \realdomain^{(N+1) \times D}$ and atoms of a system is that it facilitates fine-grained conditioning.
We can inject atom- and bond-level information by concatenating embeddings of the desired properties, such as atomic types or spatial positions, directly to the corresponding node features.
Similarly, connectivity constraints can be incorporated by augmenting the edge vectors with embeddings indicating the presence or absence of chemical bonds. 

To enable flexible conditional generation alongside \emph{de novo} tasks, we train \modelname{} across diverse conditional tasks. 
For each training sample, we randomly mask specific subsets of structural information to simulate various generation scenarios. With a probability of 0.25, we fix the positions and atom types of all MOF substructures (nodes, linkers, or solvents) except for one randomly selected component, effectively training the model for structure inpainting tasks (e.g., generating linkers within a fixed node scaffold). 
Independently, we condition on the system's chemical compositions (atom types) and bond topology with a 0.25 probability each. Whenever composition or bonds are provided, we also supply information about clustering of the fragments that identify distinct molecular fragments and components; otherwise, this clustering information is provided stochastically.
This multi-task training objective enables a single model to perform \emph{de novo} generation, conformer generation, and partial structure inpainting.
Importantly, when conditioning on structural information, the atom order index is not provided.

\section{Geometry optimization with Orb}
\label{sec:geoopt}

We perform geometry optimizations with the sum of two independent MLIPs: \texttt{\seqsplit{orb-v3-direct-20-omat}}, which was trained at the PBE level of theory and a 3-layer model with the same basic architecture, but trained to predict an additive D3 (zero) term. We run both models at \texttt{float32-highest} precision. These models are available from the \href{https://github.com/orbital-materials/orb-models}{\texttt{orb-models} package} \citep{neumannOrbFastScalable2024, rhodesOrbv3AtomisticSimulation2025}. Whilst D3 can be computed analytically, existing implementations are inefficient and create a bottleneck.

We use the Fr\'echet cell FIRE optimizer \citep{bitzek2006structural} implemented in \href{https://github.com/TorchSim/torch-sim}{\texttt{TorchSim}} \citep{cohen2025torchsim} with a $0.05$ eV/\angstrom{} max force convergence criterion and a maximum of $500$ iterations. With these settings, $84\%$ of \dbname{} and $98\%$ of QMOF converge.

\section{Three-step molecular dynamics pipeline for dynamic stability}
\label{sec:MD}

To compute the dynamic properties of MOFs in Section \ref{subsec:dynamic_mof} we used a three-step MD pipeline with similar settings to those reported by \textcite{krass2025mofsimbench} The pipeline was designed to rapidly assess the thermal stability of generated MOF samples at 300 K.

All steps used \texttt{orb-v3-con-inf-omat} combined with D3 dispersion corrections\cite{rhodesOrbv3AtomisticSimulation2025, grimme2010consistent}.
For these 1,000 samples, we used the torch D3 implementation from the \href{https://github.com/pfnet-research/torch-dftd}{\texttt{torch-dftd}} library.
The first step of the pipeline was a coarse geometry optimization using the LBFGS optimizer in ASE (maximum force threshold of 0.05 eV/\AA{}, maximum of 1,000 steps), followed by an NVT equilibration step using Langevin dynamics at 300 K (timestep 1 fs, friction coefficient of 0.01 $\mathrm{fs}^{-1}$, 1 ps duration). 
Finally, the unit cell was allowed to relax isotropically in the NPT ensemble using MTK dynamics via the \texttt{IsotropicMTKNPT} ASE implementation. This final simulation was run for 50 ps at 300 K with a timestep of 1 fs, a thermostat timescale (\textit{tdamp}) of 100 fs, and a barostat timescale (\textit{pdamp}) of 1,000 fs.
To quantify stability, the change in volume was calculated as the difference in mean cell volume between the first 10 ps and the final 10 ps of the simulation, while RMSD values were computed based on the deviation of atomic coordinates between the first and final frames.

\clearpage
\section{ORB latent embeddings of MofasaDB samples}
\label{sec:latent-embeddings}
\begin{figure}[ht!]
    \centering
    \includegraphics[width=1.0\linewidth]{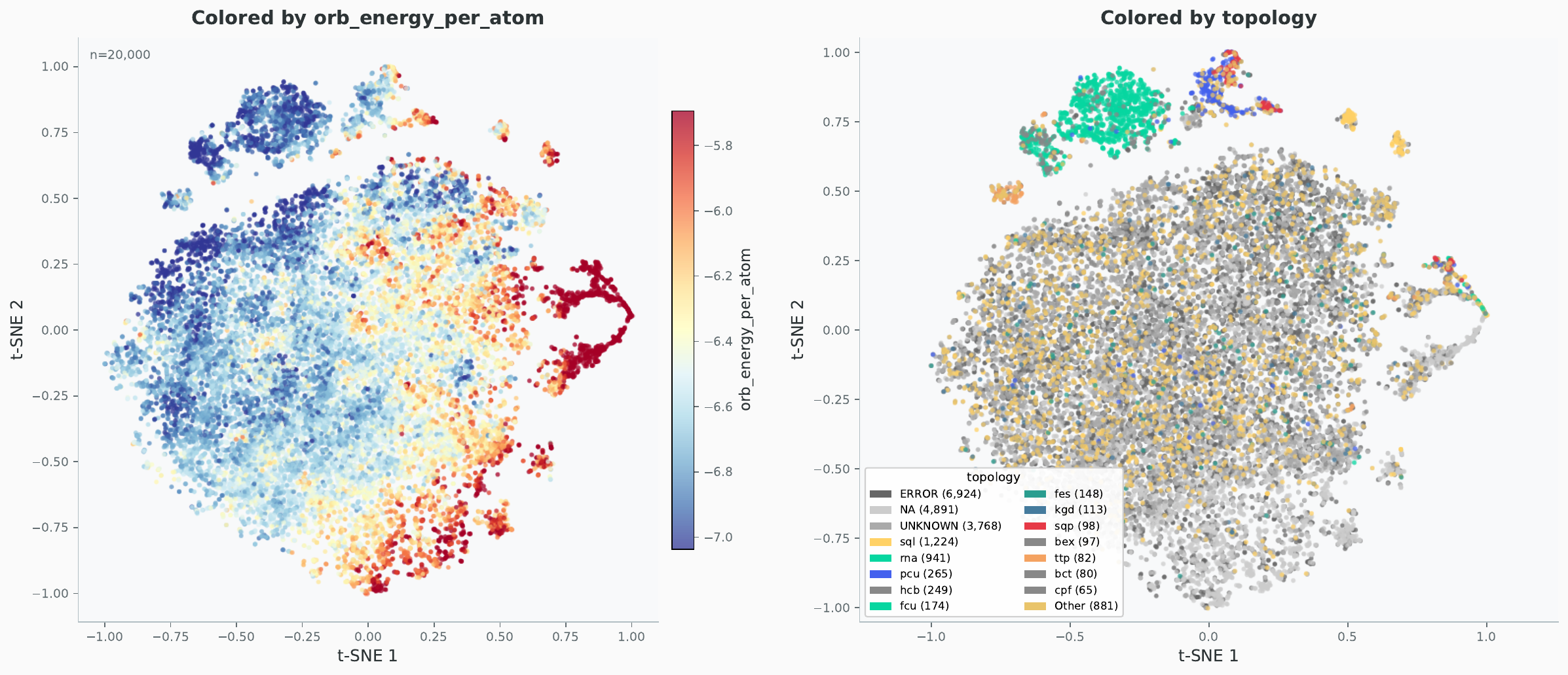}
    \caption{t-SNE plot of \texttt{orb-v3-direct-20-omat} latent embeddings (aggregated by mean) of 20,000 MofasaDB samples. Samples are colored according to their inferred \emph{total potential energy per atom} ($\text{eV} / \text{atom}$, left) and \emph{topology} (right).}
    \label{fig:tsne-latent-top-energy}

    \vspace{0.5cm}
    
    \includegraphics[width=1.0\linewidth]{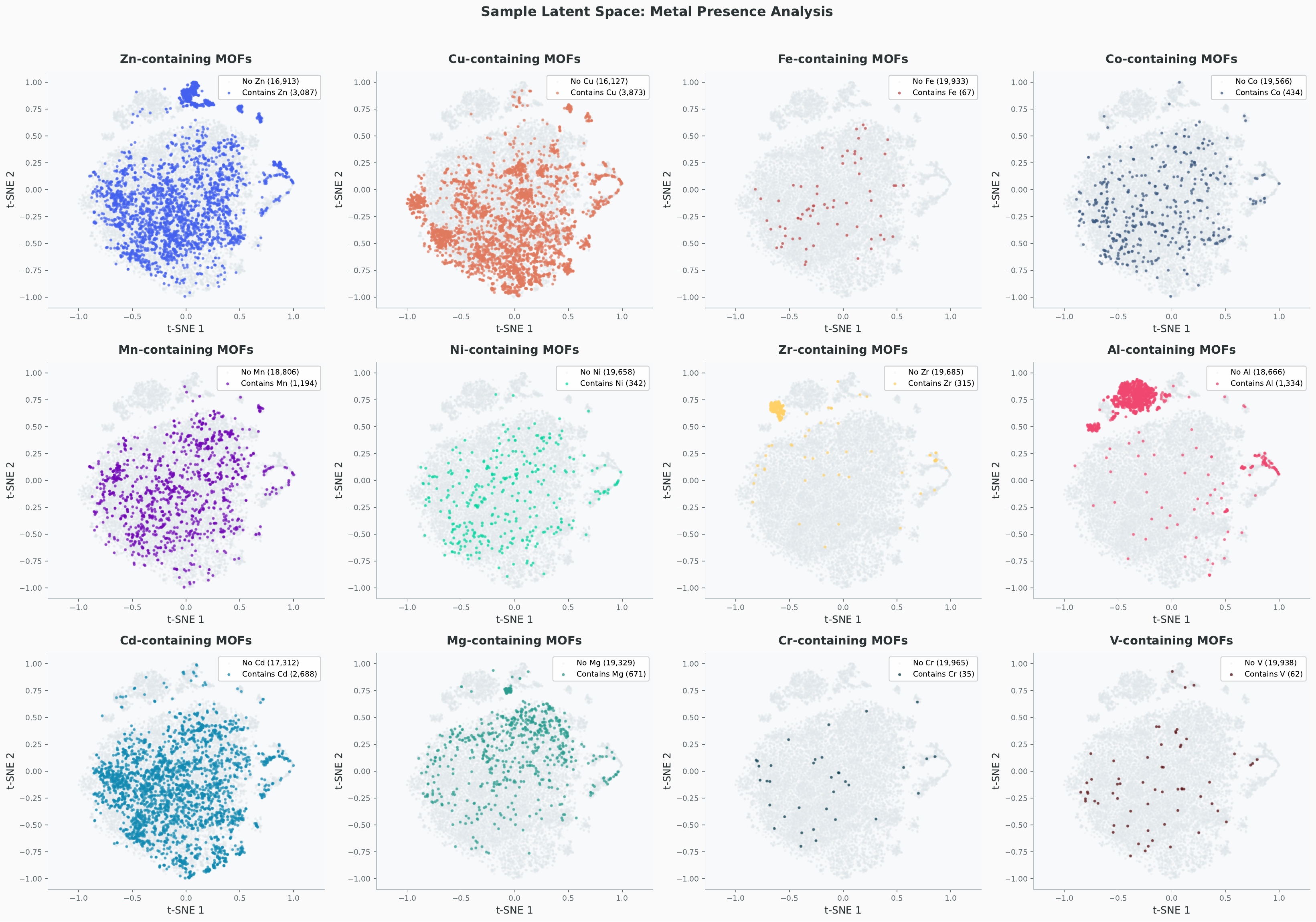}
    \caption{t-SNE plot of \texttt{orb-v3-direct-20-omat} latent embeddings (aggregated by mean) of 20,000 MofasaDB samples. Samples are colored according to the presence of specific metals in the metal nodes.}
    \label{fig:tsne-latent-metals}
\end{figure}
\clearpage

\section{Extended results}

\begin{table}[h!]
    \centering
    \begin{tabular}{@{}lrrrrr@{}}
        \toprule
        \textbf{Test (\%)} & \textbf{QMOF} & \textbf{\modelname{-opt}} & \textbf{\modelname{}} & \textbf{ADiT-QMOF} & \textbf{ADiT Joint} \\
        \midrule
        Has carbon ↑ & 100.0 & 100.0 & 98.4 & 100.0 & 100.0 \\
        Has hydrogen ↑ & 99.8 & 99.9 & 98.4 & 99.6 & 100.0 \\
        Has atomic overlap ↓ & 0.0 & 0.0 & 2.8 & 8.3 & 10.8 \\
        Has overcoord. C ↓ & 0.0 & 0.0 & 2.5 & 23.6 & 34.3 \\
        Has overcoord. N ↓ & 0.0 & 0.0 & 1.6 & 1.5 & 1.6 \\
        Has overcoord. H ↓ & 0.0 & 0.0 & 2.4 & 1.0 & 3.6 \\
        Has undercoord. C ↓ & 5.6 & 14.3 & 20.1 & 60.0 & 72.1 \\
        Has undercoord. N ↓ & 6.5 & 7.3 & 12.9 & 39.1 & 39.9 \\
        Has undercoord. rare earth ↓ & 0.0 & 0.2 & 2.1 & 0.4 & 0.8 \\
        Has metal ↑ & 100.0 & 99.2 & 97.7 & 100.0 & 99.4 \\
        Has lone molecule ↓ & 9.7 & 27.1 & 31.3 & 72.9 & 83.2 \\
        Has high charge ↓ & 1.5 & 1.6 & 3.5 & 0.9 & 2.5 \\
        Has suspicious terminal oxo ↓ & 0.0 & 0.5 & 2.5 & 2.6 & 5.8 \\
        Has undercoord. alkali ↓ & 0.1 & 1.0 & 3.0 & 1.0 & 6.4 \\
        Has geom. exposed metal ↓ & 1.7 & 3.7 & 5.3 & 7.0 & 9.6 \\
        Validity rate (all passed) ↑ & 80.1 & 59.8 & 52.9 & 15.7 & 10.2 \\
        \bottomrule
    \end{tabular}
    \caption{A breakdown of the \texttt{MOFChecker} validity criteria for \modelname{} and ADiT trained on the QMOF database. We observe that \modelname{-opt} displays fewer violations of the ``Has carbon/hydrogen/metal'' checks. However, this cannot happen since geometry optimization cannot remove or add atoms. 
    This is an artifact of \texttt{MOFChecker} failures. Whenever \texttt{MOFChecker} fails to compute for a system, all criteria default to ``False''. As \texttt{MOFChecker} failures were less frequent for optimized systems, fewer artificial violations were recorded.}
    \label{tab:mofchecker-qmof}
\end{table}

\begin{table}[h!]
    \centering
    \begin{tabular}{@{}lrrr@{}}
        \toprule
        \textbf{Test (\%)} & \textbf{Experimental} & \textbf{Mofasa-opt (Exp.)} & \textbf{Mofasa (Exp.)} \\
        \midrule
        Has carbon ↑ & 99.6 & 99.5 & 98.8 \\
        Has hydrogen ↑ & 98.2 & 97.9 & 97.2 \\
        Has atomic overlap ↓ & 0.0 & 0.0 & 1.7 \\
        Has overcoord. C ↓ & 0.0 & 0.0 & 2.4 \\
        Has overcoord. N ↓ & 0.0 & 0.0 & 1.0 \\
        Has overcoord. H ↓ & 0.0 & 0.0 & 2.1 \\
        Has undercoord. C ↓ & 3.3 & 17.1 & 24.7 \\
        Has undercoord. N ↓ & 4.1 & 6.9 & 13.5 \\
        Has undercoord. rare earth ↓ & 0.1 & 0.3 & 1.5 \\
        Has metal ↑ & 100.0 & 99.1 & 98.3 \\
        Has lone molecule ↓ & 6.5 & 18.3 & 22.6 \\
        Has high charge ↓ & 1.4 & 0.6 & 1.7 \\
        Has suspicious terminal oxo ↓ & 0.8 & 1.7 & 3.6 \\
        Has undercoord. alkali ↓ & 0.1 & 0.9 & 1.9 \\
        Has geom. exposed metal ↓ & 1.3 & 3.5 & 5.8 \\
        Validity rate (all passed) ↑ & 84.2 & 62.8 & 52.2 \\
        \bottomrule
    \end{tabular}
    \caption{A breakdown of the \texttt{MOFChecker} validity criteria for \modelname{} trained on the Experimental database. We observe that \modelname{-opt} displays fewer violations of the ``Has carbon/hydrogen/metal'' checks. However, this cannot happen since geometry optimization cannot remove or add atoms. 
    This is an artifact of \texttt{MOFChecker} failures. Whenever \texttt{MOFChecker} fails to compute for a system, all criteria default to ``False''. As \texttt{MOFChecker} failures were less frequent for optimized systems, fewer artificial violations were recorded.}
    \label{tab:mofchecker-expmof}
\end{table}

\begin{table}[h!]
    \centering
    \setlength{\tabcolsep}{3.5pt} 
    \begin{tabular}{@{}>{\footnotesize}lrrrrrrrrrrrrr@{}}
    \toprule
    \textbf{Dataset} & \textbf{N} & \textbf{E} & \textbf{EV} & \textbf{EU} & \textbf{EU-R} & \textbf{ENU} & \textbf{ENU-R} & \textbf{EVU} & \textbf{EVU-R} & \textbf{EVNU} & \textbf{EVNU-R} \\
    \midrule
    \modelname{} (QMOF) & 1k & 77.1 & 42.8 & 76.5 & 89.8 & 75.2 & 88.3 & 42.4 & 60.5 & 41.5 & 59.2 \\
    \modelname{} (QMOF) & 10k & 75.7 & 44.0 & 73.5 & 86.2 & 72.5 & 85.1 & 43.0 & 61.3 & 42.4 & 60.5 \\
    \modelname{} (QMOF) & 100k & 75.8 & 43.9 & 68.8 & 80.7 & 68.5 & 80.3 & 40.2 & 57.3 & 40.0 & 57.0 \\
    \modelname{-opt} (QMOF) & 1k & 77.0 & 51.7 & 76.4 & 89.7 & 74.8 & 87.8 & 51.3 & 73.2 & 50.6 & 72.2 \\
    \modelname{-opt} (QMOF) & 10k & 75.3 & 50.2 & 72.7 & 85.3 & 71.6 & 84.1 & 48.9 & 69.7 & 48.2 & 68.7 \\
    \modelname{-opt} (QMOF) & 100k & 75.7 & 49.8 & 68.4 & 80.3 & 68.1 & 79.9 & 45.7 & 65.1 & 45.4 & 64.8\\
    \modelname{} (Exp) & 1k & 82.3 & 46.9 & 77.6 & 84.2 & 73.6 & 79.9 & 43.1 & 54.7 & 40.1 & 50.9 \\
    \modelname{} (Exp) & 10k & 82.7 & 46.7 & 71.8 & 77.9 & 69.8 & 75.7 & 38.4 & 48.7 & 36.8 & 46.8 \\
    \modelname{} (Exp) & 100k & 82.9 & 46.5 & 67.6 & 73.3 & 66.9 & 72.6 & 35.5 & 45.0 & 34.9 & 44.4 \\
    \modelname{-opt} (Exp) & 1k & 84.6 & 57.0 & 78.4 & 85.1 & 74.0 & 80.3 & 52.0 & 66.0 & 48.7 & 61.8 \\
    \modelname{-opt} (Exp) & 10k & 83.5 & 55.0 & 72.6 & 78.7 & 70.5 & 76.5 & 46.3 & 58.8 & 44.8 & 56.8 \\
    \modelname{-opt} (Exp) & 100k & 83.4 & 56.0 & 67.6 & 73.3 & 66.9 & 72.6 & 43.7 & 55.5 & 43.2 & 54.8 \\
    \bottomrule
    \end{tabular}
    \caption{\textbf{MOFid: Existence (E), Validity (V), Novelty (N) and Uniqueness (U)} for a range of sample-sizes (N). Importantly, the real data (QMOF or ``Experimental'') does not have $100\%$ E or V, and so we also report rescaled (-R) percentages, dividing by the maximum possible score (which is a function of both dataset and metric).}
    \label{tab:vnu}
\end{table}

\begin{table}[h]
    \centering
    \begin{tabular}{@{}lrr@{}}
        \toprule
        \textbf{Transition} & \textbf{Count} & \textbf{Percentage} \\
        \midrule
        ERROR $\to$ ERROR & 107,967 & 53.5\% \\
        Valid $\to$ Valid & 38,781 & 19.2\% \\
        UNKNOWN $\to$ UNKNOWN & 30,441 & 15.1\% \\
        ERROR $\to$ UNKNOWN & 5,865 & 2.9\% \\
        UNKNOWN $\to$ ERROR & 5,731 & 2.8\% \\
        ERROR $\to$ Valid & 3,937 & 1.9\% \\
        Valid $\to$ ERROR & 3,857 & 1.9\% \\
        UNKNOWN $\to$ Valid & 2,920 & 1.4\% \\
        Valid $\to$ UNKNOWN & 2,427 & 1.2\% \\
        \bottomrule
    \end{tabular}
    \caption{Breakdown of how topologies of generated samples in MofasaDB change after geometry optimization with \texttt{orb-v3-direct-20-omat}. 
    Notably, approximately 6.4\% of samples change between Valid to ERROR/UNKNOWN states.
    This suggests that topology inferred via \texttt{MOFid} is sensitive to small geometry perturbations and should be interpreted with caution.
    }
    \label{tab:topology_change_percentage_breakdown}
\end{table}

\begin{table}[h]
    \centering
    \begin{tabular}{@{}lrr@{}}
        \toprule
        \textbf{Status} & \textbf{Count} & \textbf{Percentage} \\
        \midrule
        Same Topology & 33,762 & 87.1\% \\
        Changed Topology & 5,019 & 12.9\% \\
        \midrule
        \textbf{Total} & \textbf{38,781} & \textbf{100.0\%} \\
        \bottomrule
    \end{tabular}
    \caption{Stability of the inferred MofasaDB topologies (i.e.\ topologies that were not labeled as \texttt{UNKNOWN}, \texttt{ERROR}, or \texttt{NA}) by \texttt{MOFid} pre- and post-geometry optimization. Shows that about 12.9\% of inferred topologies changed after geometry optimization.}
    \label{tab:topology_stability}
\end{table}

\begin{table}[h]
    \centering
    \begin{tabular}{@{}lrclr@{}}
        \toprule
        \textbf{Rank} & \multicolumn{3}{c}{\textbf{Transition}} & \textbf{Count} \\
        \midrule
        1 & sqp & $\to$ & pcu & 311 \\
        2 & sql & $\to$ & pcu & 238 \\
        3 & bct & $\to$ & fcu & 184 \\
        4 & dia & $\to$ & sqp & 177 \\
        5 & cpf & $\to$ & rna & 135 \\
        6 & hcb & $\to$ & sql & 135 \\
        7 & sql & $\to$ & hcb & 122 \\
        8 & sdb & $\to$ & ttp & 85 \\
        9 & sql & $\to$ & sqp & 72 \\
        10 & pcu & $\to$ & sql & 72 \\
        11 & hcb & $\to$ & dia & 60 \\
        12 & dia & $\to$ & pcu & 57 \\
        13 & sql & $\to$ & fes & 55 \\
        14 & hcb & $\to$ & sqp & 54 \\
        15 & fes & $\to$ & sql & 51 \\
        \bottomrule
    \end{tabular}
    \caption{Top 15 topology changes in MofasaDB pre- and post-geometry optimization with \texttt{orb-v3-direct-20-omat}.}
    \label{tab:topology_top15_changes}
\end{table}

\clearpage

\section{Glossary of computed properties}
\label{sec:property-glossary}

\begin{longtable}{@{}p{5cm}lp{7.5cm}@{}}
\caption{MofasaDB Property Glossary} \label{tab:mofasadb-glossary} \\

\toprule
\textbf{Key} & \textbf{Unit} & \textbf{Description} \\
\midrule
\endfirsthead

\multicolumn{3}{c}%
{{\tablename\ \thetable{} -- continued from previous page}} \\
\toprule
\textbf{Key} & \textbf{Unit} & \textbf{Description} \\
\midrule
\endhead

\midrule
\multicolumn{3}{r}{{Continued on next page}} \\
\bottomrule
\endfoot

\bottomrule
\endlastfoot

\multicolumn{3}{l}{\textit{\textbf{Pore Geometry (Zeo++ \cite{willems2012algorithms}, default: N\textsubscript{2} probe radius of 1.86 \AA{})}}} \\
\texttt{\seqsplit{lcd}} & \AA{} & Largest Cavity Diameter \\
\texttt{\seqsplit{pld}} & \AA{} & Pore Limiting Diameter (narrowest channel point) \\
\texttt{\seqsplit{dif}} & \AA{} & Diameter of Included sphere along Free path \\
\texttt{\seqsplit{number\_of\_channels}} & — & Count of distinct connected channel systems \\
\texttt{\seqsplit{number\_of\_pockets}} & — & Count of isolated pores (inaccessible to probe) \\
\midrule

\multicolumn{3}{l}{\textit{\textbf{Volume Properties (Zeo++ \cite{willems2012algorithms}, default: N\textsubscript{2} probe radius of 1.86 \AA{})}}} \\
\texttt{\seqsplit{av\_volume\_fraction}} & — & Accessible volume fraction of unit cell \\
\texttt{\seqsplit{av\_cm3\_per\_g}} & cm³/g & Accessible pore volume per gram \\
\texttt{\seqsplit{nav\_volume\_fraction}} & — & Non-accessible (pocket) volume fraction \\
\texttt{\seqsplit{nav\_cm3\_per\_g}} & cm³/g & Non-accessible volume per gram \\
\texttt{\seqsplit{channel\_volume\_fraction}} & — & Fraction of void volume in channels \\
\texttt{\seqsplit{pocket\_volume\_fraction}} & — & Fraction of void volume in pockets \\
\midrule

\multicolumn{3}{l}{\textit{\textbf{Surface Area Properties (Zeo++ \cite{willems2012algorithms}, default: N\textsubscript{2} probe radius of 1.86 \AA{})}}} \\
\texttt{\seqsplit{asa\_m2\_per\_cm3}} & m²/cm³ & Accessible surface area per unit volume \\
\texttt{\seqsplit{asa\_m2\_per\_g}} & m²/g & Accessible surface area per gram (cf.\ BET) \\
\texttt{\seqsplit{nasa\_m2\_per\_cm3}} & m²/cm³ & Non-accessible surface area per unit volume \\
\texttt{\seqsplit{nasa\_m2\_per\_g}} & m²/g & Non-accessible surface area per gram \\
\texttt{\seqsplit{channel\_surface\_area\_fraction}} & — & Fraction of surface area in channels \\
\texttt{\seqsplit{pocket\_surface\_area\_fraction}} & — & Fraction of surface area in pockets \\
\midrule

\multicolumn{3}{l}{\textit{\textbf{Crystal Symmetry (Pymatgen \cite{ongPythonMaterialsGenomics2013})}}} \\
\texttt{\seqsplit{spacegroup}} & str & Crystal system from space group analysis at symprec=0.01 (e.g., ``cubic'') \\
\texttt{\seqsplit{spacegroup\_v2}} & str & Crystal system from space group analysis at symprec=0.1 (more tolerant) \\
\texttt{\seqsplit{symprec\_0.01/pointgroup}} & str & Point group symbol (Hermann-Mauguin notation) \\
\texttt{\seqsplit{symprec\_0.01/spacegroup}} & str & Space group symbol (Hermann-Mauguin notation) \\
\texttt{\seqsplit{symprec\_0.01/spacegroup\_number}} & int & International Tables space group number (1-230) \\
\texttt{\seqsplit{symprec\_0.01/spacegroup\_crystal}} & str & Crystal system name \\
\texttt{\seqsplit{symprec\_0.1/pointgroup}} & str & Point group symbol (at looser tolerance) \\
\texttt{\seqsplit{symprec\_0.1/spacegroup}} & str & Space group symbol (at looser tolerance) \\
\texttt{\seqsplit{symprec\_0.1/spacegroup\_number}} & int & Space group number (at looser tolerance) \\
\texttt{\seqsplit{symprec\_0.1/spacegroup\_crystal}} & str & Crystal system name (at looser tolerance) \\
\midrule

\multicolumn{3}{l}{\textit{\textbf{ORB Properties \cite{rhodesOrbv3AtomisticSimulation2025}}}} \\
\texttt{\seqsplit{orb\_energy\_per\_atom}} & eV/atom & ORB-predicted total potential energy per atom \\
\texttt{\seqsplit{orb\_max\_force}} & eV/\AA{} & Maximum atomic force magnitude \\
\texttt{\seqsplit{orb\_latent\_\{0-4\}\_\{pool\}}} & — & GNN latent embeddings (dim=256); \texttt{pool} $\in$ \{\texttt{graph}, \texttt{nodes\_and\_bridges}, \texttt{linkers}, \texttt{bound\_solvent}, \texttt{free\_solvent}\} \\
\midrule

\multicolumn{3}{l}{\textit{\textbf{MOF Fragment Properties}}} \\
\texttt{\seqsplit{\{component\}\_formulas}} & List[str] & Chemical formulas per fragment; \texttt{component} $\in$ \{\texttt{nodes\_and\_bridges}, \texttt{linkers}, \texttt{bound\_solvent}, \texttt{free\_solvent}\} \\
\texttt{\seqsplit{linkers\_smiles}} & List[str] & Full SMILES strings for each linker fragment \\
\texttt{\seqsplit{linkers\_simple\_smiles}} & List[str] & Simplified scaffold SMILES (no stereochemistry) \\
\midrule

\multicolumn{3}{l}{\textit{\textbf{Linker Molecular Descriptors \cite{landrumRDKitOpensourceCheminformatics2025}}}} \\
\texttt{\seqsplit{linkers\_smiles\_used}} & List[str] & Which SMILES string was successfully parsed for each linker (original, fixed, or simple) \\
\texttt{\seqsplit{linkers\_smiles\_standardized}} & List[str] & Neutralized, canonical tautomer SMILES \\
\texttt{\seqsplit{linkers\_morgan\_ecfp\{4,6\}[\_std]}} & — & Morgan fingerprints (2048-bit); \texttt{\_std} = standardized \\
\texttt{\seqsplit{linkers\_morgan\_count\_sum}} & List[int] & Sum of Morgan fingerprint bit counts (molecular complexity proxy) \\
\texttt{\seqsplit{linkers\_morgan\_count\_sum\_max}} & List[int] & Maximum count in Morgan fingerprint (indicates highly represented substructures) \\
\texttt{\seqsplit{linkers\_morgan\_count\_sum\_std}} & List[int] & Sum of counts for standardized fingerprints \\
\texttt{\seqsplit{linkers\_morgan\_count\_sum\_max\_std}} & List[int] & Maximum count for standardized fingerprints \\
\texttt{\seqsplit{linkers\_rotatable\_bonds}} & List[int] & Number of rotatable bonds per linker (flexibility metric) \\
\texttt{\seqsplit{linkers\_ring\_count}} & List[int] & Number of rings per linker \\
\texttt{\seqsplit{linkers\_coordination\_site\_count}} & List[int] & Total number of potential metal coordination sites per linker \\
\texttt{\seqsplit{linkers\_coordination\_site\_breakdown}} & List[Dict] & Breakdown by coordination site type \\
\texttt{\seqsplit{linkers\_carboxylate\_count}} & List[int] & Number of carboxylate groups (-COO$^{-}$/-COOH) \\
\texttt{\seqsplit{linkers\_pyridine\_count}} & List[int] & Number of aromatic nitrogen sites \\
\texttt{\seqsplit{linkers\_imidazole\_n\_count}} & List[int] & Number of imidazole/triazole NH groups \\
\texttt{\seqsplit{linkers\_primary\_amine\_count}} & List[int] & Number of primary amine groups (-NH$_2$) \\
\texttt{\seqsplit{linkers\_secondary\_amine\_count}} & List[int] & Number of secondary amine groups (-NH-) \\
\texttt{\seqsplit{linkers\_tertiary\_amine\_count}} & List[int] & Number of tertiary amine groups (-N$<$)\\
\texttt{\seqsplit{linkers\_phosphonate\_count}} & List[int] & Number of phosphonate groups \\
\texttt{\seqsplit{linkers\_sulfonate\_count}} & List[int] & Number of sulfonate groups \\
\texttt{\seqsplit{linkers\_phenolic\_oh\_count}} & List[int] & Number of phenolic hydroxyl groups \\
\texttt{\seqsplit{linkers\_alcoholic\_oh\_count}} & List[int] & Number of alcoholic hydroxyl groups \\
\texttt{\seqsplit{linkers\_thiol\_count}} & List[int] & Number of thiol groups (-SH) \\
\texttt{\seqsplit{linkers\_nitrile\_count}} & List[int] & Number of nitrile groups (-C$\equiv$N) \\
\midrule

\multicolumn{3}{l}{\textit{\textbf{Validation Metrics}}} \\
\texttt{\seqsplit{no\_atom\_too\_close}} & bool & True if all interatomic distances are reasonable \\
\texttt{\seqsplit{smact\_valid}} & bool & True if SMACT charge-balance check passes \\
\midrule

\multicolumn{3}{l}{\textit{\textbf{MOFChecker Properties \cite{jin2025mofchecker}: used for validation}}} \\
\texttt{\seqsplit{mofchecker\_valid}} & bool & Overall validity flag (True if passes all checks) \\
\texttt{\seqsplit{mofchecker\_no\_carbon}} & bool & True if structure contains no carbon atoms \\
\texttt{\seqsplit{mofchecker\_no\_hydrogen}} & bool & True if structure contains no hydrogen atoms \\
\texttt{\seqsplit{mofchecker\_no\_metal}} & bool & True if structure contains no metal atoms \\
\texttt{\seqsplit{mofchecker\_has\_atomic\_overlaps}} & bool & True if atoms are too close (clashing) \\
\texttt{\seqsplit{mofchecker\_has\_lone\_molecule}} & bool & True if structure contains disconnected fragments \\
\texttt{\seqsplit{mofchecker\_has\_overcoordinated\_c}} & bool & True if any carbon has too many bonds \\
\texttt{\seqsplit{mofchecker\_has\_overcoordinated\_n}} & bool & True if any nitrogen has too many bonds \\
\texttt{\seqsplit{mofchecker\_has\_overcoordinated\_h}} & bool & True if any hydrogen has too many bonds \\
\texttt{\seqsplit{mofchecker\_has\_undercoordinated\_c}} & bool & True if any carbon has too few bonds \\
\texttt{\seqsplit{mofchecker\_has\_undercoordinated\_n}} & bool & True if any nitrogen has too few bonds \\
\texttt{\seqsplit{mofchecker\_has\_undercoordinated\_rare\_earth}} & bool & True if any rare earth metal is undercoordinated \\
\texttt{\seqsplit{mofchecker\_has\_undercoordinated\_alkali\_alkaline}} & bool & True if alkali/alkaline earth metal is undercoordinated \\
\texttt{\seqsplit{mofchecker\_has\_suspicious\_terminal\_oxo}} & bool & True if incorrect terminal oxo groups exist \\
\texttt{\seqsplit{mofchecker\_has\_geometrically\_exposed\_metal}} & bool & True if metal has unusual coordination geometry \\
\texttt{\seqsplit{mofchecker\_has\_high\_charges}} & bool & True if computed partial charges are unusually high \\
\midrule
\multicolumn{3}{l}{\textit{\textbf{MOFChecker Properties \cite{jin2025mofchecker}: not used for validation}}} \\
\texttt{\seqsplit{mofchecker\_has\_oms}} & bool & True if structure has Open Metal Sites \\
\texttt{\seqsplit{mofchecker\_has\_3d\_connected\_graph}} & bool & True if framework is 3D-connected \\
\texttt{\seqsplit{mofchecker\_graph\_hash}} & str & Hash of the full structure graph (atoms + bonds) \\
\texttt{\seqsplit{mofchecker\_undecorated\_graph\_hash}} & str & Hash of graph with hydrogen atoms removed \\
\texttt{\seqsplit{mofchecker\_decorated\_scaffold\_hash}} & str & Hash of framework scaffold with decorations \\
\texttt{\seqsplit{mofchecker\_undecorated\_scaffold\_hash}} & str & Hash of bare framework scaffold \\
\texttt{\seqsplit{mofchecker\_symmetry\_hash}} & str & Hash encoding symmetry information \\
\midrule

\multicolumn{3}{l}{\textit{\textbf{MOFid Properties \cite{buciorIdentificationSchemesMetal2019}}}} \\
\texttt{\seqsplit{mofid}} & str & Full MOFid identifier string. Format: \{nodes\}.\{linkers\} MOFid-v1.\{topology\}.cat\{n\}. Value is ``UNKNOWN'' if MOFid could not be computed. \\
\texttt{\seqsplit{mofkey}} & str & MOFKey identifier (a hash-based representation). Format: \{hash\}.\{topology\}.MOFkey-v1.\{short\_code\}. Value is ``UNKNOWN'' if MOFKey could not be computed. \\
\texttt{\seqsplit{nodes}} & str & Concatenated SMILES strings of all distinct metal nodes (.-separated). Value is ``UNKNOWN'' if not available. \\
\texttt{\seqsplit{linkers}} & str & Concatenated SMILES strings of all distinct organic linkers (.-separated). Value is ``UNKNOWN'' if not available. \\
\texttt{\seqsplit{num\_distinct\_nodes}} & int & Number of chemically distinct metal node types in the MOF \\
\texttt{\seqsplit{num\_distinct\_linkers}} & int & Number of chemically distinct organic linker types in the MOF \\
\texttt{\seqsplit{topology}} & str & Three-letter RCSR topology code (e.g., ``pcu'', ``dia'', ``fcu''). Value is ``UNKNOWN'' if topology could not be determined. \\
\texttt{\seqsplit{topology\_v2}} & str & Alternative topology assignment (may differ from primary if ambiguous) \\
\texttt{\seqsplit{catenation}} & int & Catenation number (degree of interpenetration). 0 = non-catenated, n = n-fold catenated \\

\end{longtable}

\end{appendices}
\end{document}